\documentclass[final,5p,times,twocolumn,authoryear]{elsarticle}

\usepackage{amssymb}
\usepackage{lipsum}
\usepackage{amsmath,amssymb,amsfonts}
\usepackage{hyperref}
\usepackage{algorithmic}
\usepackage{graphicx}
\usepackage{textcomp}
\usepackage{graphicx}
\usepackage{color}
\usepackage{tabularray}
\usepackage{amsmath}
\usepackage{graphicx}
\usepackage{colortbl}
\setcitestyle{square}
\usepackage{comment}
\usepackage{xcolor}
\usepackage{float}
\usepackage{multirow}
\usepackage{fancybox}
\usepackage[round]{natbib}
\usepackage{url}
\usepackage{adjustbox, booktabs, multirow, multicol, amssymb, url}
\usepackage{fontawesome}  

\usepackage{color}

\definecolor{darkgreen}{rgb}{0.0,0.5,0.0}

\journal{Neural Networks}

\newcommand{\methodname}{TextEconomizer}
\newcommand{\vectorname}{Kizuki}

\begin{document}

\begin{frontmatter}



\title{TextEconomizer: Enhancing Lossy Text Compression with Denoising Transformers and Entropy Coding}


\author[1]{Mahbub E Sobhani\fnref{github}}
\author[1]{Anika Tasnim Rodela}
\author[2]{Chowdhury Mofizur Rahman}
\author[3]{Dewan Md. Farid}
\author[2]{Swakkhar Shatabda\corref{cor1}}

\cortext[cor1]{Corresponding author. \href{mailto:swakkhar.shatabda@bracu.ac.bd}{swakkhar.shatabda@bracu.ac.bd}}

\fntext[github]{\small{\href{https://github.com/mahbubhimel/TextEconomizer}{\faGithub\enspace github.com/mahbubhimel/TextEconomizer}}}
\affiliation[1]{organization={United International University}, Department of Computer Science and Engineering,
                city={Dhaka},
                country={Bangladesh}}
\affiliation[2]{organization={BRAC University}, Department of Computer Science and Engineering,
                city={Dhaka},
                country={Bangladesh}}
\affiliation[3]{organization={Southeast University}, Department of Computer Science and Engineering, 
                city={Dhaka},
                country={Bangladesh}}

\begin{abstract}
Lossy text compression reduces data size while preserving core meaning, making it well-suited for tasks like summarization, automated analysis, and digital archives where exact fidelity is less critical. Despite the dominance of transformer-based models in language modeling, the integration of context vectors and lossless entropy coding into Sequence-to-Sequence (Seq2Seq) text generation remains underexplored. A key challenge lies in identifying the most informative context vectors from the encoder output and incorporating entropy coding into the transformer framework to enhance storage efficiency while maintaining high-quality outputs, even in the presence of noisy text. Previous studies have primarily focused on near-lossless token generation, often overlooking space efficiency. In this paper, we introduce {\methodname}, an encoder-decoder framework paired with a transformer neural network. This framework utilizes its latent representation to reduce variable-sized inputs by 50\% to 80\%, without prior knowledge of dataset dimensions. Our model achieves competitive compression ratios by incorporating entropy coding, while delivering near-perfect text quality, as assessed by Bilingual Evaluation Understudy (BLEU), Recall-Oriented Understudy for Gisting Evaluation (ROUGE), Metric for Evaluation of Translation with Explicit ORdering (METEOR), and semantic similarity scores. Notably, {\methodname} operates with approximately 153 times fewer parameters than comparable models, achieving a compression ratio of 5.39$\times$ without sacrificing semantic quality. Additionally, we evaluate our framework by implementing a Long Short-Term Memory (LSTM)-based autoencoder, commonly used in image compression, and by integrating advanced modules within the transformer architecture as alternatives to conventional techniques. Our autoencoder achieves a state-of-the-art compression ratio of 67$\times$ with 196 times fewer parameters, while our modified transformer, LLaMAFormer, achieves a 263-fold reduction in parameters compared to ICAE while maintaining competitive text quality. The {\methodname} framework significantly surpasses existing transformer-based models in balancing memory efficiency and high-fidelity outputs, marking a breakthrough in lossy compression with optimal space utilization. All the codes and data are available at GitHub:  \url{https://github.com/mahbubhimel/TextEconomizer}
\end{abstract}




\end{frontmatter}




\section{Introduction}
Text compression is a key component of data compression that reduces the volume of textual data while retaining its core informational content. This process can be either lossless, which allows for complete data recovery, or lossy, which sacrifices certain details to achieve higher compression ratios. The exponential growth of digital information has created significant challenges in storage and transmission efficiency \citep{textVermont}, particularly in contexts where exact textual reproduction is not mandatory. Lossy text compression offers numerous practical applications, providing effective solutions for non-critical documents and archiving \cite{palaniappan2007lossy}, where exact preservation of every detail is unnecessary. This approach is useful for internal reports, outdated document versions, corporate archives \cite{niu2023corporate}, and educational institutions, where retaining key information is crucial, but minor errors or formatting loss are not detrimental to usability. Libraries, digital archives \cite{leggett2020digitization}, and universities can compress extensive collections, such as research papers or public domain books, saving storage space while ensuring continued accessibility. Furthermore, industries like chatbots \cite{rahimi2023compression}, email archiving \cite{bhuvaneswari2023exploring}, and web search engines \cite{williams2012compression} benefit from lossy text processing, enabling quicker retrieval and facilitating communication \cite{ECIR2014}. Thus, lossy text compression emerges as a valuable and efficient tool for managing large-scale textual data, optimizing both storage and retrieval performance.

Image compression harnessing neural networks, especially through transformers, has gained prominence \citep{liu2023learned}, meanwhile myriad approaches have been discovered to reduce textual volume while preserving salient information. Existing research underscores the efficacy of transformer-based models, including BERT \citep{li2023crossword}, LLaMA \citep{valmeekam2023llmzip}, and ALBERT \citep{li2021text}, LSTM \citep{prato-etal-2019-towards}, in maintaining contextual integrity during decompression across diverse linguistic landscapes. In particular, architectural modifications such as the shared encoder have shown strong performance \citep{li2020explicit}. Furthermore, chasing the trend of fine-tuning domain-specific pre-trained models and incorporating the Low-Rank Adaptation (LoRA) \citep{hu2021lora} technique has also resulted in praiseworthy results \citep{ge2023context}. Cross-lingual augmentation strategies enhance the capabilities of transformer models for languages with diverse resource availability, an aspect that has not been investigated in the work of \citep{mao2022trace}. Notwithstanding these advancements, recent studies have encountered limitations. The study conducted by \citep{huang2023approximating} leveraged the operational procedure of arithmetic coding and incorporated it with Generative Pre-trained Transformer (GPT) for lossless text compression, lacking a quest for harnessing general-purpose compressors for an extensive comparison of compression ratios. Additionally, it is important to note that non-autoregressive decoding may not flawlessly recover the original text, and iterative inspection of meaning preservation might be time-consuming \citep{ge2022lossless}. \citep{ge2023context} employed a LoRA-configured Llama-2-7b for context compression, albeit with augmented parameters, exacerbating the already prodigious parameter count characteristic of contemporary LLMs. Meanwhile, \citep{qin2023nugget} executed an autoencoding task generating dynamic text segments with residual connections, achieving a mere tenfold compression—a ratio deemed insufficient. Furthermore, \citep{wang2021tsdae} provided an autoencoding model trained to reconstruct input texts by means of a combination of token embeddings as a bottleneck, a strategy susceptible to overfitting, while the compression ratio \textit{ r} is lower. Strengthening the fixed-size bottleneck remains a pivotal challenge for Transformer-based large language models (LLMs) due to their intrinsic self-attention mechanism. While prior research (\citep{rae2019, malireddy2020scar, wang2021tsdae}) has explored text compression in LLMs, they frequently grapple with the challenge of mitigating memory intricacy. Despite their capacity for fixed-size latent spaces, LSTM-based autoencoders frequently produce inadequate results in autoregressive decoding tasks. Using these orthodox approaches, we tackle the issue of memory complexity from an alternative standpoint: employing lossy text compression.

Several challenges have been identified in the domain of text autoencoding for English, notably regarding the consequences of computationally efficient models that can achieve high compression ratios while preserving the integrity of the original text. Moreover, there is a need for standardized benchmarks to evaluate the performance of such models across diverse corpora. This study intends to surmount these challenges by introducing a novel encoder-decoder framework tailored for English autoencoding tasks that can be seamlessly incorporated with different neural network architectures, including LSTMs \cite{hochreiter1997long} and both traditional and modern transformer components (e.g., Rotary Positional Encoding (RoPE) \cite{su2024roformer}, Sigmoid-Weighted Linear Unit (SwiGLU) \cite{swiGLU}, Root Mean Square Layer Normalization (RMSNorm) \cite{zhang2019root}), thereby establishing a substantial foundation for future research. To this end, we introduce {\methodname}, a monolingual transformer-based method that demonstrates competitive performance with significantly fewer parameters than existing baselines. Additionally, we have incorporated entropy coding on the top-k most salient context vectors, known as {\vectorname} vectors, to further enhance compression efficiency, and concurrently investigated the integration of trendy neural network components to optimize our framework's performance. Building on this, our framework employs a sophisticated noisy text approach to forge a wide range of text distortions, enabling the model to learn effective denoising strategies. The encoder processes the noisy input using positional encoding and self-attention mechanisms to generate continuous representations of {\vectorname} vectors, which are then utilized by the decoder to reconstruct the original text. Through persistent bottleneck refinement and the selection of paramount context vectors, the model achieves remarkable efficiency in compression while keeping high fidelity in text reconstruction. The contribution of our study is summarized below: 

\begin{itemize}
    \item We design a pragmatic text noise process that encompasses diverse real-world distortions, enabling our unified encoder–decoder framework to learn robust text-denoising behavior across different neural architectures.

    \item To demonstrate our framework's capabilities, we present two lightweight transformer models. {\methodname} delivers competitive performance with $\sim$153$\times$ fewer parameters than ICAE, NUGGET, T5-Small, and LSTM baselines, while LLaMAFormer incorporates RoPE, SwiGLU, and RMSNorm to improve text-generation quality with even fewer parameters.

    \item We evaluate distilled context vectors for capturing linguistic patterns and enhancing compression utilizing entropy coding, enhancing memory efficiency. Additionally, we evaluate how varying context vector proportions affect key metrics.

    \item We investigate how training corpus size and attention mechanisms affect the autoencoder’s ability to accurately reconstruct text, providing insights into its potential for neural decompression.

    \item We demonstrate that a monolingual autoencoder can achieve state-of-the-art memory efficiency with negligible text-quality degradation, extending the relevance of autoencoding from image compression to text-based applications.
    
\end{itemize}

The following sections of the paper are organized as follows: Sect. \ref{relatedWorks} surveys existing research in text compression, underlining transformers, and neural network-based techniques along with their inherent limitations. Sect. \ref{proposed_method} outlines the methodology and architecture of {\methodname}, a monolingual transformer-based model for efficient English autoencoding, covering its encoder, kizuki selector, decoder, and entropy coding. Sect. \ref{exp_analysis} presents a comprehensive empirical evaluation, detailing datasets, preprocessing, augmentation, corpus statistics, experimental settings, evaluation metrics, and an ablation study, with results that benchmark {\methodname}’s performance against state-of-the-art methods and provide qualitative insights. Finally, Sect. \ref{conclusion_futureWork} summarizes our findings, discusses the implications for memory-efficient NLP, and outlines promising avenues for future inquiry.

\section{Related Work}
\label{relatedWorks}
The domain of text compression has gained widespread attention of research, attracting significant interest and contributing to the development of innovative methodologies and datasets. Our extensive study aims to summarize key findings and showcase the evolving landscape of text compression methodologies across different language contexts, including lossy \citep{li2023crossword, li2020explicit, li2021text, ge2022lossless} and lossless \citep{valmeekam2023llmzip, huang2023approximating, mao2022trace} techniques.

\subsection{Transformer-based methods}
The advent of text compression has seen the employment of transformer-based methods where GPT, Llama, Bidirectional and Auto-Regressive Transformers (BART), and even a single-layer transformer have been utilized. Among lossy techniques, \citep{li2023crossword} introduced an innovative method to compress English text by masking less important words and then restoring them using a Transformer-based model, while \citep{rae2019} refined the Transformer architecture \citep{vaswani2017attention} by employing a compressive-memory-based approach that consolidates past activations rather than discarding them. In addition, \cite{zhang2025l3tc} proposed a new text compression method called L3TC, which combines learning-based probabilistic modeling with efficient architecture by using an Receptance Weighted Key Value (RWKV) language model \citep{peng-etal-2023-rwkv} instead of a traditional transformer, along with an outlier-aware tokenizer and high-rank reparameterization to achieve high compression rates (upto 50\%); however, the study lacks empirical validation on compression ratio over different batch sizes, might struggle with low-frequency words, and has not been thoroughly tested for resilience with large-scale data. Furthermore, \cite{li2021text} proposed an approach aimed at extracting the core information of the input text by enhancing the text encoding of transformer models with text compression through two distinct strategies: Explicit Text Compression (ETC) and Implicit Text Compression (ITC), the latter producing compressed text features without generating an explicit text sequence by using a non-autoregressive decoder. Moreover, \cite{elias2025multitokvariablelengthtokenizationefficient} introduced MultiTok, a variable-length tokenization method motivated by the Lempel-Ziv-Welch (LZW) algorithm, which dynamically compresses duplicative phrases into multi-word tokens to achieve up to 2.5$\times$ faster training with 30\% less data while maintaining performance comparable to BERT and GPT-2 standards. In a related vein, \citep{qin2023nugget} introduced an interesting methodology that leverages the BART \citep{lewis2019bart} encoder to produce highly significant dynamic text segments, termed “NUGGET,” by distilling the logits through a feed-forward network. Subsequently, \cite{aghanya2025lossy} introduced Error-Bounded Predictive Coding (EPC), a lossy text compression method that employs a masked language model for decompression. It masks reliably predicted tokens and stores small residuals for incorrect predictions. This method achieves 1.1 bits per character (bpc) and a BERTScore of 99.97 using a 125M RoBERTa-base model, allowing precise control over data rate and fidelity. Additionally, our study has revealed that lossless text compression techniques have yielded exceptional results by integrating diverse approaches with transformers; for example, \citep{valmeekam2023llmzip} proposed a method that uses the LLM to predict the next token in a text sequence based on a window of past tokens, and \citep{huang2023approximating} introduced a technique that employs the GPT model to calculate probability distributions for each token, thereby representing the entire text with a single number, while \citep{deletang2023language} compared predictive models and lossless compressors, recommending the use of large self-supervised language models for compression. Arithmetic-coding–based transformer compressors include \cite{mao2025lossless}, who achieved a 23$\times$ compression ratio using large language models (LLMs) with 1B to 14B parameters for accurate predictions. They applied an arithmetic coding mechanism to convert these prediction probabilities into a minimal bitstream. \cite{zhang2025test} introduced the Test-Time Steering method, which combines a pre-trained language model with a Naive Bayes \cite{sobhani2025adaptive} compressor using a Weighted Product of Experts. This method achieved 8.12\% compression with a 3-billion-parameter language model by optimizing scalar weights based on a single input sample, using the resulting probability distribution for lossless compression with an arithmetic coder. Among context-compression-based methods, \cite{chevalier-etal-2023-adapting} introduced AutoCompressors—a novel strategy that adapts pre-trained language models such as OPT \citep{zhang_opt} and Llama-2—to compress long contexts into consolidated summary vectors that are generated consecutively and accumulated across segments, thus enabling efficient processing of long documents while retaining essential information. In a similar context, \citep{ge2023context} proposed an innovative approach that leverages the capabilities of the Llama model \citep{touvron2023llama} to generate pertinent memory slots through the teacher-forcing mechanism for both autoencoding and machine translation tasks. Furthermore, \cite{mu2024learning} introduced a technique known as “gisting,” which involves training a language model to condense prompts into smaller sets of “gist” tokens, achieving up to 26 times compression, a 40\% reduction in floating-point operations (FLOPs), and a 4.2\% improvement in processing speed, while maintaining good output quality with minimal loss. \cite{qin-etal-2024-dodo} introduced DODO, a dynamic context compression method for decoder-only language models that reduces self-attention costs by depicting text with a variable number of hidden states per layer, thereby maintaining strong performance in language modeling, question-answering, and summarization, and serving as both an autoregressive model and a context compressor for subsequent tasks.

\subsection{RNN-based Autoencoder methods}
Within LSTM-based autoencoding tasks, \citep{malireddy2020scar} introduced an indicator vector to signify the presence or omission of each word, thereby filtering less pertinent words. Concurrently, \citep{tissier2019near} proposed an autoencoder-based model that condenses real-valued embedding vectors into fixed-length binary representations, while \citep{acharya2019online} enhanced storage efficiency by decomposing the embedding layer through matrix factorization and employing lower-rank matrices. Expanding these autoencoding frameworks, \citep{prato-etal-2019-towards} developed a recursive autoencoder that encapsulates entire sentences into single embeddings to optimize decodability, rather than merely enriching latent representations. In contrast, \citep{joint_source_channel} introduced an LSTM-based architecture for joint source-channel coding, embedding sentences into semantic spaces to preserve information and outperform traditional methods under low bit allocations and error rates. Parallelly, in lossless compression-focused research, \citep{goyal2018deepzip} leveraged recurrent neural networks with arithmetic coding for lossless compression, achieving near-ideal results on synthetic datasets and surpassing Gzip on real-world data. Further advancing neural compression techniques, \citep{9272105} proposed an innovative text compression method combining deep neural networks with Canonical Huffman Codes, reducing message size by up to 30\% compared to conventional algorithms. Expanding the application scope of such frameworks, \citep{geleta2023deep} introduced a variational autoencoder (VAE)-based model for population genetics, compressing single-nucleotide polymorphism (SNP) data losslessly while generating synthetic genomes that preserve genetic diversity patterns.

\subsection{Denoising Autoencoder-based methods}
Recent research has demonstrated the versatility of denoising autoencoders (DAEs) in enhancing natural language processing tasks through diverse architectural and methodological innovations. \cite{shen2020educating} proposed extending adversarial autoencoders (AAEs) with a denoising objective to improve latent space geometry, introducing the Denoising Adversarial Autoencoder (DAAE), which reconstructs original sentences from perturbed versions to encourage the encoder to map semantically similar texts closer in the latent space. \cite{lopez-avila-suarez-paniagua-2023-combining} developed a three-phase framework combining denoising autoencoders with contrastive learning, alongside a novel data augmentation method for supervised contrastive learning to mitigate dataset imbalance; their approach adapts pre-trained Transformer models for classification tasks, enhancing performance by aligning representations with the target data distribution before fine-tuning. Another innovative approach, SUNDAE, introduced by \cite{savinov2022stepunrolleddenoisingautoencoderstext}, employs step-unrolled denoising autoencoders to iteratively refine token sequences, achieving state-of-the-art results in machine translation and competitive performance in unconditional language modeling without relying on autoregressive architectures. Furthermore, \cite{freitag2018unsupervised}  explored unsupervised text generation from structured data by framing structured inputs as noisy targets, leveraging sequence-to-sequence DAEs to reconstruct fluent sentences; their method, which introduces controlled noise during training, proved effective across multiple languages and domains without requiring labeled data.

Transformer-based techniques, though superior in autoregressive decoding evaluation, encounter memory constraints due to a large number of parameters. Despite these successes, the use of latent representations within transformers for text compression is still largely unexplored, and there has been limited research on filtering context vectors before passing them to the decoder. To the best of our knowledge, no existing work proposes a denoising transformer framework for autoencoding that eliminates noise, utilizes a filtered latent space based on contextual coherence to generate high-fidelity outputs, and incorporates lossless entropy coding to enhance compression. This distinctive combination establishes the novelty of our approach to addressing these gaps.

\section{Proposed Method}
\label{proposed_method}
\begin{figure*}[!htb] 
  \centering
  \includegraphics[width=0.8\textwidth]{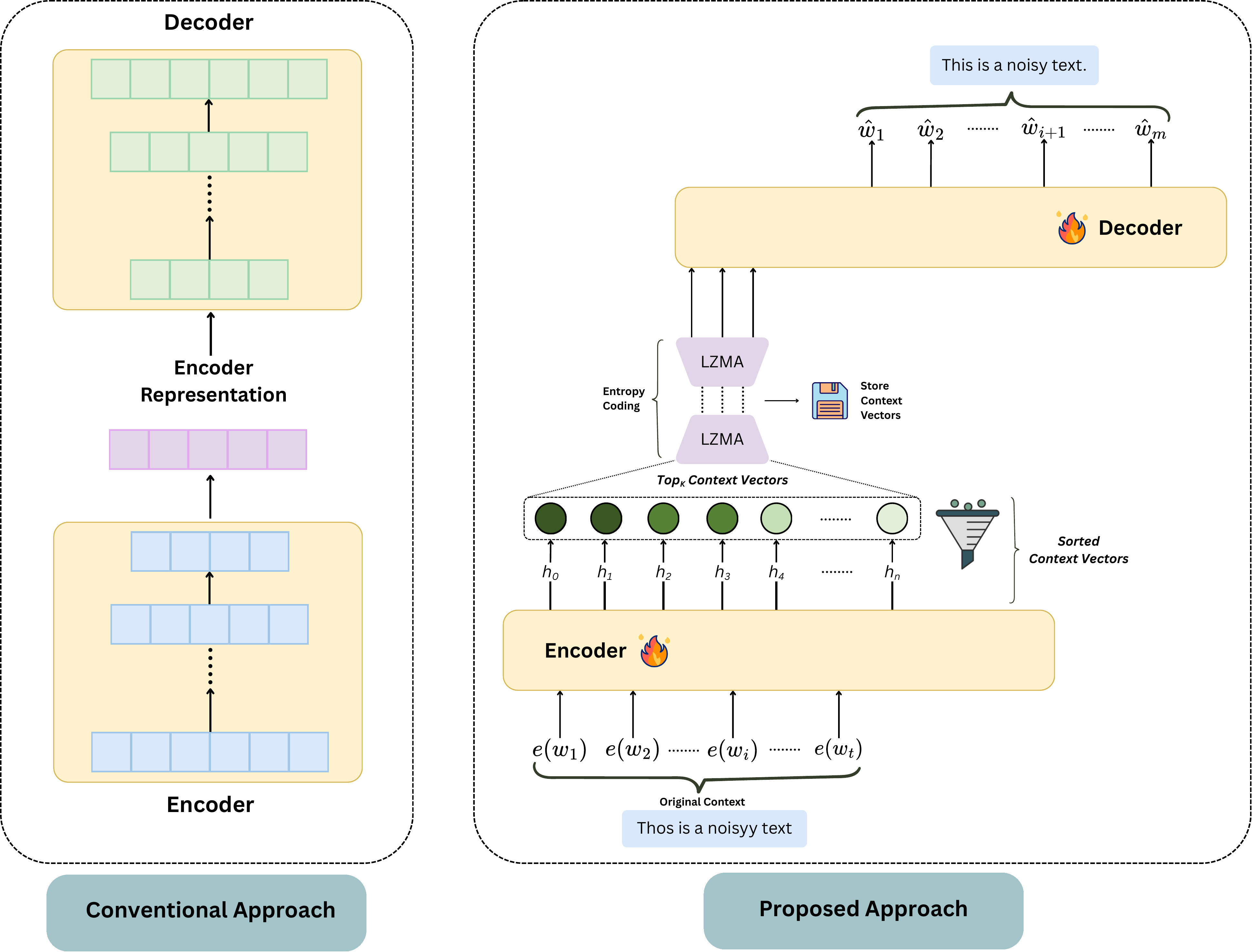} 
  \caption{\textbf{(Left)} Illustration of a conventional transformer encoder-decoder architecture where the latent representation retains the full dimensionality of the input sequence, often resulting in redundant calculations. \textbf{(Right)}  Our proposed framework is designed for efficient text reconstruction. The model encodes noisy input text into a sequence of context vectors ($h_0, ..., h_n$). A probabilistic sorting prioritizes these vectors based on information density, followed by a $Top_k$ selection strategy to isolate the most salient {\vectorname} vectors. These vectors are then compressed using Entropy Coding (LZMA) for further memory efficiency before being decoded to generate the corrected output ($\hat{w}$). This pipeline effectively reduces memory overhead without sacrificing reconstruction quality, enabling more efficient downstream decoding and transmission.}
  \label{fig:textecono}
\end{figure*}

We propose a seq2seq encoder-decoder framework for English autoencoding that seamlessly integrates any neural network-based method while confirming memory efficiency and high-quality output. We assessed our framework through a tripartite approach. We first implemented our transformer-based model, optimized to select the most salient context vectors during training, termed {\methodname}. Second, we replaced conventional transformer components with Rotary Positional Encoding, RMSNorm, and SwiGLU activation functions to create LLaMAFormer, which similarly targets the most critical context vectors. Third, we constructed an LSTM-based autoencoder to assess the performance of sequential methods in autoencoding tasks. All three approaches leverage our innovative pragmatic text-denoising process to guide the autoencoding workflow. The complete English autoencoding framework is illustrated in Fig. \ref{fig:textecono}. The subsequent sections delineate the problem formulation and the components of our proposed TextEconomizer.

\subsection{Problem Formulation \& Overview}
Consider input sentence $S = \{S_1, S_2, \ldots, S_{N-1},S_{N}\}$, where $N$ denotes the word count. The noise injection process ($\aleph(\cdot)$) adroitly introduces strategic realistic noise through a meticulous automated supervision protocol.  Subsequently, we scrutinize the juxtaposition of token pairs, $X_{I} = \{x_1, x_2, \ldots, x_{n-1}, x_{n}\}$ and $Y_{I} = \{y_1, y_2, \ldots, y_{k-1}, y_{k}\}$, where $X_{I}$ epitomizes the input sequence with multifarious noise types, while $Y_{I}$ embodies the pristine target sequence. The erroneous sentence $X_I$ is tokenized using a pre-trained tokenizer $\tau(\cdot)$. This process produces a tokenized representation of $X$, denoted as $X = \{x_{\tau_1}, x_{\tau_2}, \ldots, x_{\tau_{n-1}}, x_{\tau_n}\}$, where $x_{\tau_i}$ corresponds to the numerical value of the $i$-th token if the word is present in the vocabulary. If the word is not present, an $\langle \text{unk} \rangle$ token is used instead. The tokenized sequence is then passed to the encoder $E(\cdot)$, which processes it to generate context vectors for each subword in the sentence. This raises an important query: are these context vectors truly indispensable? To address this question, we conducted a meticulous filtering process, denoted by the $\triangledown$, wherein only the most salient context vectors; referred to as {\vectorname} vectors, deemed the most significant are allowed to proceed to subsequent stages. {\vectorname}s are collectively represented as $\mathcal{Z}$. This latent representation $\mathcal{Z}$ is further subjected to compaction through the Lempel–Ziv–Markov compressor $\mathcal{LC}(\cdot)$ to facilitate parsimonious storage on a hard disk, whereupon memory-related computations are conducted. The Lempel–Ziv–Markov decompressor $\mathcal{LD}(\cdot)$ reconstitutes the compressed representation to a form indistinguishable from $\mathcal{Z}$. The decoder $D(\cdot)$ uses this latent representation, incorporating the teacher forcing concept, to remove the noise and synthesize the correct sentence non-autoregressively during training. This pipeline exhibits seamless integration capabilities with any Transformer-based encoder-decoder architecture. The entire procedure can be encapsulated in the following mathematical formulation:
\vspace{6pt}

\begin{equation}
    \hat{Y} = D((\mathcal{LD}(\mathcal{LC}(\nabla(E(\tau([X_I]), W^E)))), D_{out}^{t-1}), W^D)
    \label{eq:problem_formulation}
\end{equation}

\vspace{4pt}

\subsection{TextEconomizer}

Our model is a tweaked version of the \cite{vaswani2017attention} style sequence-to-sequence (seq2seq) transformer architecture, specifically designed for English text compression and autoencoding tasks. The preference for a vanilla transformer architecture was inspired by the success of recent transformer-based models, including BART, LLaMA, and GPT, which have achieved state-of-the-art results on autoencoding benchmarks such as WMT19 and the PWC test set. Our model includes a stack of six encoder and decoder blocks, where individual blocks contain self-attention mechanisms and feedforward neural networks. Below, we provide detailed descriptions of the encoder and decoder blocks.

\subsection{Encoder}
We transform each sentence into a sequence of tokens $ X = \{x_1, x_2, \dots, x_n\} $ and assign discrete values to each subword. Here, $ n $ is the sequence length. To ensure consistent input dimensions for every input sequence, we incorporate padding into $X$. Subsequently, each token $ x_i $ is passed through an embedding layer $E$, which transforms the discrete values into continuous vectors represented by a trainable matrix ${E}_{x_i} = {Embedding}(x_i) $. Through backpropagation during training, these embeddings are fine-tuned to minimize the loss. To preserve the positional order of the tokens in a sentence, we incorporate positional encoding ${PE}$ for each token $ x_i $, denoted as $ {PE}_{x_i} $. This positional encoding is added to the embedding, resulting in a composite embedding $ {Z}_{x_i} = {E}_{x_i} + {PE}_{x_i} $. The composite embedding is then fed into a stack of $ {K} $ identical encoder layers, which are built upon two main components: a position-wise feed-forward network and a multi-head self-attention mechanism. The self-attention mechanism is the key segment of the Transformer architecture. It computes attention scores between all pairs of tokens in the input sequence $ {X} $ using three learnable matrices: query ($ {Q} $), key ($ {K} $), and value ($ {V} $). These matrices capture the contextual dependencies between tokens in the sequence. Specifically, the self-attention mechanism calculates a weighted sum of the hidden states for each token, where the weights are based on the similarity between the token and all other tokens in the sequence. This allows the encoder to focus on the most relevant parts of the input sequence for each token, taking into account the context in which it appears. The mathematical expression of self-attention is defined as follows \cite{vaswani2017attention}: 
\begin{equation}
    \resizebox{.70\hsize}{!}{${Attention}({Q}, {K}, {V}) = {softmax}(\frac{{QK}^T}{\sqrt{{d_k}}})  V$}
    \label{eq:self_attention}
\end{equation}
Subsequently, the dimension of the keys denoted as ${d}_k$, is used to scale the resultant scores for standardizing overall variance and stabilizing gradients. To enhance the model's ability to capture multiple relationships and increase its expressive power, the transformer employs multi-head self-attention. Each head independently computes self-attention, allowing the model to focus on different parts of the input sequence simultaneously. The outputs of all heads are concatenated and linearly transformed to produce the final output. The self-attention for each head is calculated as:

\begin{equation}
    \text{Head}_i = \text{Attention}({Q}_i, {K}_i, {V}_i)
\end{equation}

where ${Q}_i = {Z}{W}^Q_i$, ${K}_i = {Z}{W}^K_i$, and ${V}_i = {Z}{W}^V_i$ represent the query, key, and value projections for the $i$-th head. The final multi-head self-attention output is calculated as:
\begin{equation}
{MultiHead}({Q}, {K}, {V}) = {Concat}({Head_1}, {Head_2}, \ldots, {Head_h}) {W}^O
\end{equation}
Here, ${W}^O$ is a learnable weight matrix that merges the outputs of all heads. Following the multi-head self-attention mechanism, the output is passed through a position-wise feed-forward neural network (FFN). This FFN applies a non-linear transformation to the output of the self-attention mechanism, enhancing the model's capacity to capture complex relationships between tokens. Specifically, the FFN consists of two linear transformations with a rectified linear unit (ReLU) activation function in between, introducing non-linearity and enriching the representation of the input sequence. The output of each encoder layer, which integrates contextual information from the multi-head self-attention mechanism and the non-linear transformations of the FFN, is sequentially passed to the next of the $K$ identical encoder layers. This stacking of layers enables the transformer encoder to progressively construct a rich, contextualized representation of the input sequence ${X}$, capturing both local and global dependencies that are critical for understanding the meaning and context of each token within the sentence. Ultimately, the final encoder layer produces a sequence of hidden states that effectively encapsulates the semantic and contextual information of each token in the input sentence.

\subsection{Kizuki Selector}
To optimize the decoding process and reduce redundant computations in the cross-attention layer of the decoder, we leverage the fact that contextual token embeddings intrinsically capture the semantics of their surrounding text. By analyzing the attention scores computed between all pairs of token embeddings, we observe that some embeddings have very low attention scores, indicating their minimal contribution to the decoding process. To address this inefficiency, we propose a selective mechanism for identifying the most informative context vectors. Specifically, given a set of contextual token embeddings ${Z} = \{z_1, z_2, \dots, z_n\}$, where $z_i \in \mathbb{R}^d$ represents the $i$-th embedding and $n$ is the total number of tokens. We first quantify the relevance of each token using a lightweight scoring function, $\phi: \mathbb{R}^d \to \mathbb{R}$. This function is a learnable Multi-Layer Perceptron (MLP) that maps an embedding to a scalar logit representing its raw saliency. This is computed over RMS-normalized context vectors to ensure training stability and invariance to scaling artifacts. We compute a probability distribution over these embeddings using a softmax function with a dynamic temperature parameter ${T}$ \cite{guo2017calibration}. To focus strictly on semantic content, special tokens (e.g., \texttt{<bos>}, \texttt{<eos>}) are excluded from this calculation. The probability $p_i$ for the $i$-th token is given by:

\begin{equation}
p_i = \text{Softmax}(z_i; T) = \frac{\exp(\phi(z_i) / T)}{\sum_{j \in \mathcal{S}} \exp(\phi(z_j) / T)},
\end{equation}

where $\mathcal{S}$ represents the set of indices corresponding to the sentence's middle tokens. The temperature $T$ is dynamically modulated as a function of the sequence length $n$ to balance exploration and exploitation in token selection. For short sequences with limited context, a larger $T$ is used to smooth the distribution, preventing overconfidence in single tokens. Conversely, for longer sequences, $T$ approaches zero to sharpen the distribution, approximating an argmax operation to distinctively isolate high-impact tokens. From this probability distribution, we select the top $k$ most informative embeddings, referred to as \textbf{\vectorname}, denoted as ${\mathcal{E}} = \{\mathcal{E}_1, \mathcal{E}_2, \dots, \mathcal{E}_k\}$, where $k = \lceil r \cdot n \rceil$ and $r$ is the reduction ratio ($0 < r \leq 1$). The selection procedure can be represented as:

\begin{equation}
\mathcal{E} = \mathop{\text{topK}}_{z_i \in Z} \left( p_i \right),
\end{equation}

This operation retains exactly the $k$ most semantically significant embeddings. These {\vectorname} vectors are then further compressed leveraging the Lempel-Ziv-Markov chain Algorithm (LZMA) \citep{ziv1978compression} algorithm to enhance memory efficiency, which works by using a sliding window to find repeated patterns in the data and substitute them with shorter codes, then applying a range encoder that calculates how often things appear and gives shorter codes to the common stuff and eventually producing a compressed representation that can be reversed back to its actual form. The compressed representation is then reversed back to the {\vectorname} vectors, which are then projected to form the Key ($K$) and Value ($V$) vectors in the decoder's cross-attention layer $K, V = \mathcal{E}$. This approach ensures that only the most semantically significant embeddings are included in the decoding process, thus reducing computational redundancy while preserving essential contextual information. By focusing on high-impact tokens, {\vectorname} enhances both computational efficiency and the model's ability to capture crucial contextual relationships during decoding.

\subsection{Decoder}
The target sequence, denoted as ${Y} = \{y_1, y_2, \dots, y_m\}$, where $m$ portrays the sequence length, is first processed through an embedding layer. This layer transforms discrete token values into continuous vector representations using the mapping ${E}_{y_i} = \text{Embed}(y_i)$. To keep positional information within the sequence, a positional encoding ${PE}_{y_i}$ is incorporated into the token embedding, resulting in the combined representation ${Z}_{y_i} = {E}_{y_i} + {PE}_{y_i}$. The combined embeddings ${Z}_{y}$ are thereafter passed through ${L}$ identical decoder layers. Each decoder layer consists of two primary components: a masked multi-head self-attention mechanism and a position-wise feed-forward network. The masking in the decoder assures that each token can only attend to prior tokens, stopping access to future tokens during the generation process. The masked multi-head self-attention mechanism follows the same computational formulation as the encoder's self-attention mechanism \eqref{eq:self_attention}. This sub-layer enables the decoder to grasp dependencies among tokens within the output sequence, enriching contextual understanding. Following the self-attention mechanism, the decoder applies a cross-attention layer to compute attention over the {\vectorname} vectors ${\mathcal{E}} = \{\mathcal{E}_1, \mathcal{E}_2, \dots, \mathcal{E}_k\}$. This multi-head attention layer allows the decoder to attend to the top-$k$ contextualized representations of the input sequence while processing the output sequence, thereby facilitating adequate alignment between input and output tokens. Subsequently, a position-wise feed-forward network is applied to enhance the learned representation through a non-linear transformation. This transformation includes residual connections and layer normalization, consistent with the procedure utilized in the encoder's feed-forward network. These operations distill the learned representation and contribute to the generation of the final output sequence. During training, the decoder employs a technique known as teacher forcing, wherein the ground-truth token from the earlier time step is fed as input at each decoding step. In contrast, during inference, the decoder produces the output sequence sequentially, predicting the most probable token at each step while considering both the previously generated tokens and the {\vectorname} vectors. The processed representation undergoes ${L}$ identical decoder layers, ultimately producing the refined target sequence.

\section{Experimental Analysis}
\label{exp_analysis}
\subsection{Dataset}
We outline the datasets used in our experiments, highlighting their fundamental characteristics. These datasets have been meticulously curated to uphold semantic coherence and syntactic diversity, ensuring their exemplary quality for research purposes. For the autoencoding task, we included only the columns containing English text in our experiments.

\textbf{WMT14 English-French (EN-FR)} \citep{bojar2014findings} The WMT14 dataset is a widely-used benchmark for machine translation, originally containing approximately 36 million sentence pairs from sources such as Europarl v7, Common Crawl, and News Commentary. Due to computational limitations, we used a subset of the EN-FR corpus: 600K training pairs, 3K validation pairs, and 3K test pairs. Despite its reduced size, this subset remains effective for evaluating translation quality.

\textbf{WMT19 English-Chinese (EN-ZH)} \citep{barrault-etal-2019-findings} The WMT19 dataset provides a diverse EN-ZH parallel corpus. We selected a subset of 600K training pairs and 3.98K test pairs. This subset ensures a robust evaluation of translation performance while maintaining topic coverage.

\textbf{Prompt-With-Context (PWC)} \citep{ge2023context} The PWC dataset assesses models in context-dependent settings. It has approximately 242K training instances and 18.1K test instances. To enhance robustness testing, we systematically added noise to the original sentences, simulating real-world noisy input scenarios.

\textbf{BookCorpus} \citep{zhu2015aligning} The BookCorpus dataset comprises text from books, offering diverse linguistic styles and contexts. We introduced synthetic noise to simulate corrupted data and partitioned the dataset into 1M training pairs and 20K test pairs. This setup evaluates model generalization in autoencoding tasks.

\subsection{Data Preprocessing}
From the WMT19 corpus, we extracted the English sentences from the Chinese-to-English pairs. We also obtained English text from WMT14's English-to-French pairs. For the PwC dataset, we isolated the answer column. Since BookCorpus is monolingual, no extraction was needed. This systematic approach produced an English-centric corpus across all datasets. In our text preprocessing, we delineate a comprehensive character set containing 80 frequently occurring English characters, denoted as $DC = \{DC_1, DC_2, \ldots, DC_{80}\}$. This set is augmented by 14 frequently occurring punctuation marks in English, represented as $PM = \{PM_1, PM_2, \ldots, PM_{14}\}$, and a space character $SP$. The resulting amalgamated set of 95 English characters is defined as $C = \{DC + PM + SP\} = \{C_1, C_2, \ldots, C_{95}\}$. Subsequently, we consider each sentence in our corpus, denoted as $S = \{S_1, S_2, \ldots, S_N\}$, where $N$ illustrates the total number of characters in the sentence. We then employ an iterative technique, examining each character $S_i \in S$ and systematically eradicating any character not present in our predefined character set $C$. This meticulous preprocessing assures a standardized and sophisticated textual dataset for subsequent neural network processing.

\subsection{Data Augmentation}
\label{data_aug}
We introduce a sophisticated noise injection technique to introduce controlled linguistic variability. This process operates on the premise that each sentence constitutes a finite set of lexical units, denoted as $S = \{W_1, W_2, \ldots, W_{N-1}, W_N\}$ where $N$ is the length of the sentence such that $N \in Z^{+}$. The proposed noisy text corruption process aims to forge an altered version of the input sentence S that preserves its semantic meaning while incorporating realistic linguistic perturbations. The process commences with identifying named entities $NE = \{e_1, e_2, \ldots, e_m\}$ utilizing a Named Entity Recognition (NER) model. Subsequently, the sentence undergoes identifying part-of-speech (POS) tags $T = \{(W_i, t_i), (W_2, t_2), \ldots, (W_N, t_N)\}$ are assigned. Auxiliary verbs $V_a\subseteq S$  are probabilistically omitted with a probability $P_{aux}$, constrained to words where $W_i \in V_a$ and $t_i \in V^*$, where $V^*$ encompasses all verb forms. Consequently, we obtain the modified set $\tilde{S} = \{\tilde{W}_1, \tilde{W}_2, \ldots, \tilde{W}_{M-1}, \tilde{W}_M\}$. The corruption process is controlled by a normally distributed corruption probability $p_c \sim \mathcal{N} (\mu_p = 0.6, \sigma^{2}_p (\sigma = 0.1))$, which is bounded by a maximum corruption threshold $p_{max} = 0.5$ to ensure the degree of alteration remains within acceptable limits. Given $p_c$, we determine the number of words to be corrupted as $k = \left\lceil M \times p_c \right\rceil$. These words are carefully chosen to avoid consecutive corruptions, preserving the sentence structure.  For each chosen word $\tilde{W}_i$, the corruption method is selected based on POS tags and whether $\tilde{W}_i \in NE$ or critical nouns and verbs $C = \{c_1, c_2, \ldots, c_l\} \subset \tilde{S}$. Specifically, if $\tilde{W}_i \notin NE$ and $t_i \in \{P_{nouns}, P_{verbs}, P_{adj}\}$, a contextual synonym $W^{'}_i$ is generated using a masked language model with a $p_{mlm} = 0.5$ probability, unless $W^{'}_i = \tilde{W}_i$. Alternatively, spelling augmentation is propagated $p_{spelling} = 0.3$ of the time, while random word substitution is employed in the remaining $p_{sub} = 0.2$ of cases. Words that do not fall into the categories of nouns, verbs, or adjectives are subjected to typographical alterations through character interchanges or replacements. Punctuation corruption is then applied with a $p_{punc} = 0.2$ probability per word, eradicating existing punctuation or introducing new punctuation marks. This meticulous process pinnacles in a corrupted sentence $S^{'}$ that mimics natural language errors, echoing common mistakes while preserving the essential meaning of the original sentence $S$.


\subsection{Corpus Statistics}
In our curated English corpus, characterized by intended noise injection, we have selected $\approx$ 246K, 600K, 600K, and 1M source-target pairs from the PwC, WMT14, WMT19, and BookCorpus datasets, respectively, due to resource limitations. Within these pairs, the source sentences undergo a meticulous text noise technique, while the target sentences serve as pristine, noise-free counterparts. To achieve this, we systematically introduced perturbations into each sentence, as outlined in Subsection \ref{data_aug}. Furthermore, the corpus exhibits the following linguistic statistics: PwC demonstrates a minimum of 1, a maximum of 180, and a mean of 35.52 words per sentence; WMT14 ranges from 1 to 72 words, averaging 21.68; WMT19 spans from 1 to 137 words, with a mean of 12.67; and BookCorpus holds sentences varying from 2 to 150 words, averaging 15.13 words per sentence.

\subsection{Hyperparameters}
The hidden dimension is kept as 512 through all the encoder and decoder layers for maintaining consistency. To maintain the model's depth and capacity the number of neurons is kept at 2048 for the feed-forward layer and a 0.1 dropout ratio is applied to prevent overfitting. To maintain the efficient computation and non-linearity over the network we have incorporated ReLU. The model went through 50 epochs with $5 \times 10^{-5}$ learning rate incorporating AdamW optimizer. We incorporated the categorical cross-entropy loss function for the optimization process, which leads the model towards desired translations. 

\subsection{Performance Evaluation}
We evaluated our model's performance in the autoencoding task by removing noise from text and calculating BERTScore \eqref{eq:f1}, BLEU \eqref{eq:sacreBLEU}, ROUGE, METEOR \eqref{eq:meteor_single},and Perplexity scores.

\subsubsection*{BERTScore. \cite{bert-score}}
The BERTScore calculates the semantic similarity of two pieces of text by calculating the cosine similarity of their embedding tokens. This metric outputs precision, recall, and f1 score, and their equations are as follows:
    \begin{equation}
        Recall_{\mathrm{BERT}}= \frac{1}{N}\times \sum_{1}^{N}\left( \frac{1}{|x|} \sum_{x_i \in x} \max _{\hat{x}_j \in \hat{x}} \mathbf{x}_i^{\top} \hat{\mathbf{x}}_j \right)
        \label{eq:recall}
    \end{equation}
    \begin{equation}
        Precision_{\mathrm{BERT}}= \frac{1}{N}\times \sum_{1}^{N}\left( \frac{1}{|\hat{x}|} \sum_{\hat{x}_j \in \hat{x}} \max _{x_i \in x} \mathbf{x}_i^{\top} \hat{\mathbf{x}}_j \right)
        \label{eq:precision}
    \end{equation}
    \begin{equation}
        F1_{BERT} = \frac{1}{N}\times \sum_{1}^{N} (2\times \frac{P_{BERT} \times R_{BERT}}{P_{BERT}+R_{BERT}})\
        \label{eq:f1}
    \end{equation}
where $\mathbf{x}_i^{\top} \hat{\mathbf{x}}_j$ is the cosine similarity between two pieces of text and N is the total number of sentence pairs. 
Recall\_BERT matches the scores of reference and candidate text's each token whereas, Precision\_BERT calculates a matching score between candidate and reference text and F1\_BERT is the harmonic mean of Recall\_BERT and Precision\_BERT.

\subsubsection*{BLEU. \cite{post-2018-call}}
The BLEU metric estimates the quality of candidate text by assigning precision scores to n-grams and comparing them with one or more reference texts. Scores range from 0 and 100, where a higher score denotes better results. The mathematical formula for BLEU is as follows:
    \begin{equation}
        \text{BLEU} = \text{BP} \times e^{\sum_{n=1}^{N}(w_{n}\cdot logp_{n})}
        \label{eq:sacreBLEU}
    \end{equation}

Here, The Brevity Penalty (BP) punishes shorter predictions. $N$ is the maximum n-gram length. $w_{n}$ are the weights for n-gram precision, and log$p_{n}$ is the logarithm of n-gram precision in the candidate text.

\subsubsection*{ROUGE \citep{lin2004rouge}}
ROUGE (Recall-Oriented Understudy for Gisting Evaluation) is a widely used metric for evaluating text generation tasks, particularly in summarization and machine translation.
\begin{itemize}
    \item ROUGE-1 measures unigram (single-word) overlap between the generated and reference translations, providing a basic lexical similarity assessment.
    \item ROUGE-2 focuses on bigram (two-word sequence) overlap, capturing short-range contextual consistency.
    \item ROUGE-L evaluates the Longest Common Subsequence (LCS), which considers word order and fluency, making it more aligned with human judgment than simple n-gram overlap metrics.
\end{itemize}
By analyzing different granularities of text similarity, ROUGE provides a comprehensive measure of text generation quality.
\begin{figure}[t]
    \centering
    \includegraphics[width=0.5\textwidth]{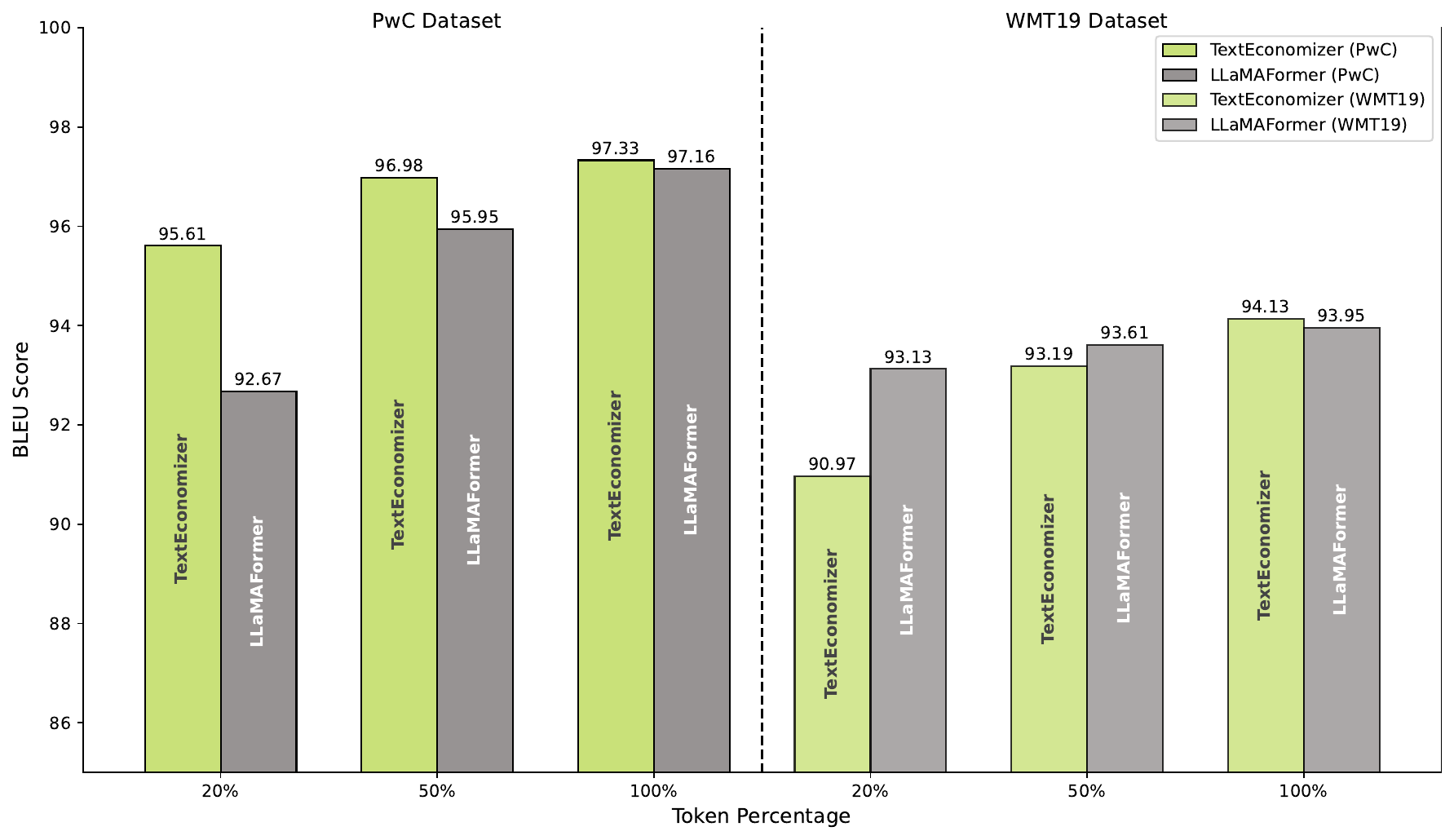}
    \caption{Comparative analysis of BLEU scores for {\methodname} and LLaMAFormer across different token utilization tiers (20\%, 50\%, 100\%) on the PwC and WMT19 datasets, where higher scores signify improved translation fidelity.}
    \label{fig:ablation_bleu}
\end{figure}


\begin{figure}[t]
    \centering
    \includegraphics[width=0.5\textwidth]{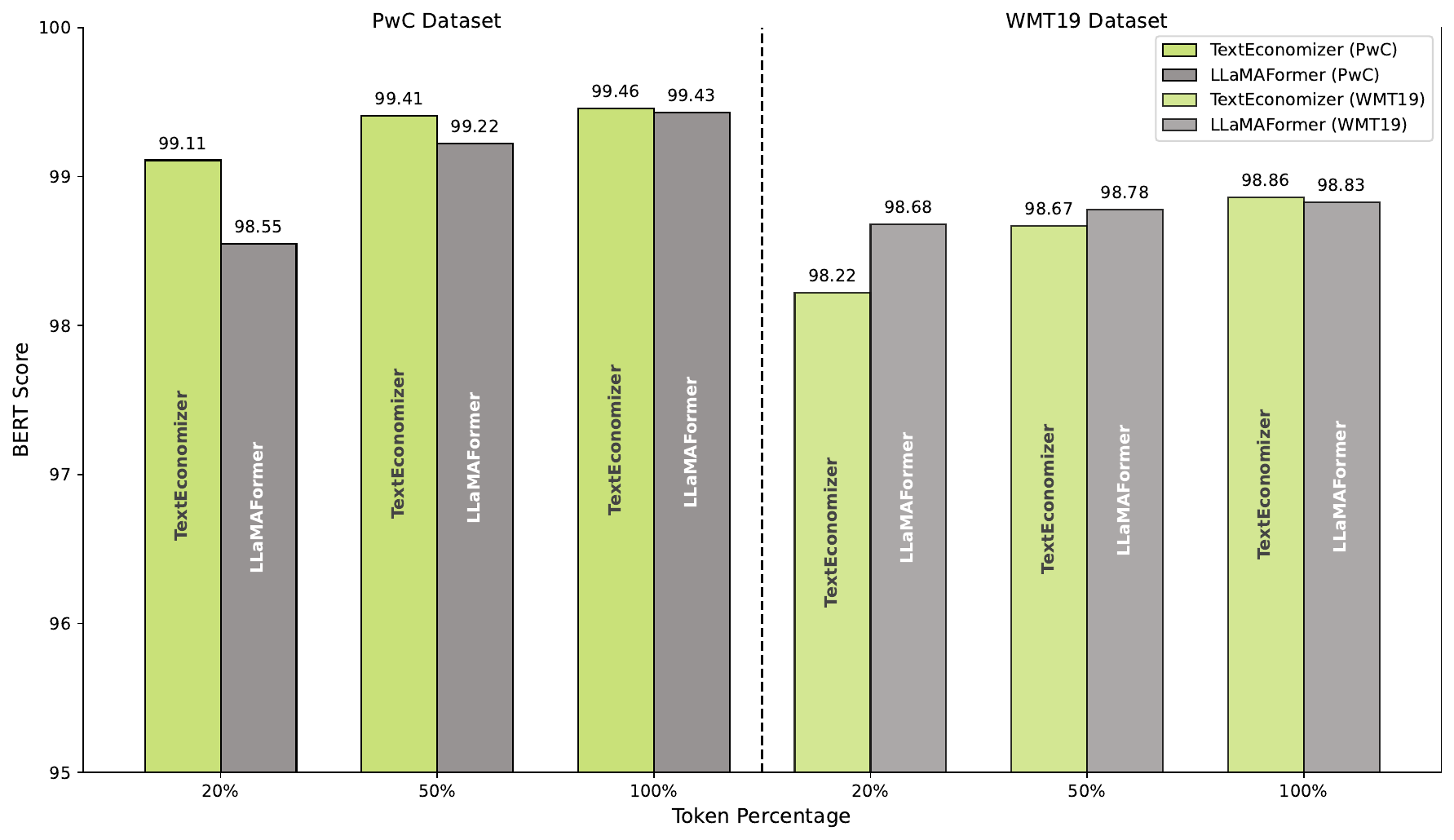}
    \caption{A juxtaposition of BERT scores between {\methodname} and LLaMAFormer indicates subtle disparities in semantic fidelity across token usage levels (20\%, 50\%, 100\%) on the PwC and WMT19 datasets, with higher values denoting refined contextual acuity.}
    \label{fig:ablation_bert}
\end{figure}
\begin{figure}[t]
    \centering
    \includegraphics[width=0.5\textwidth]{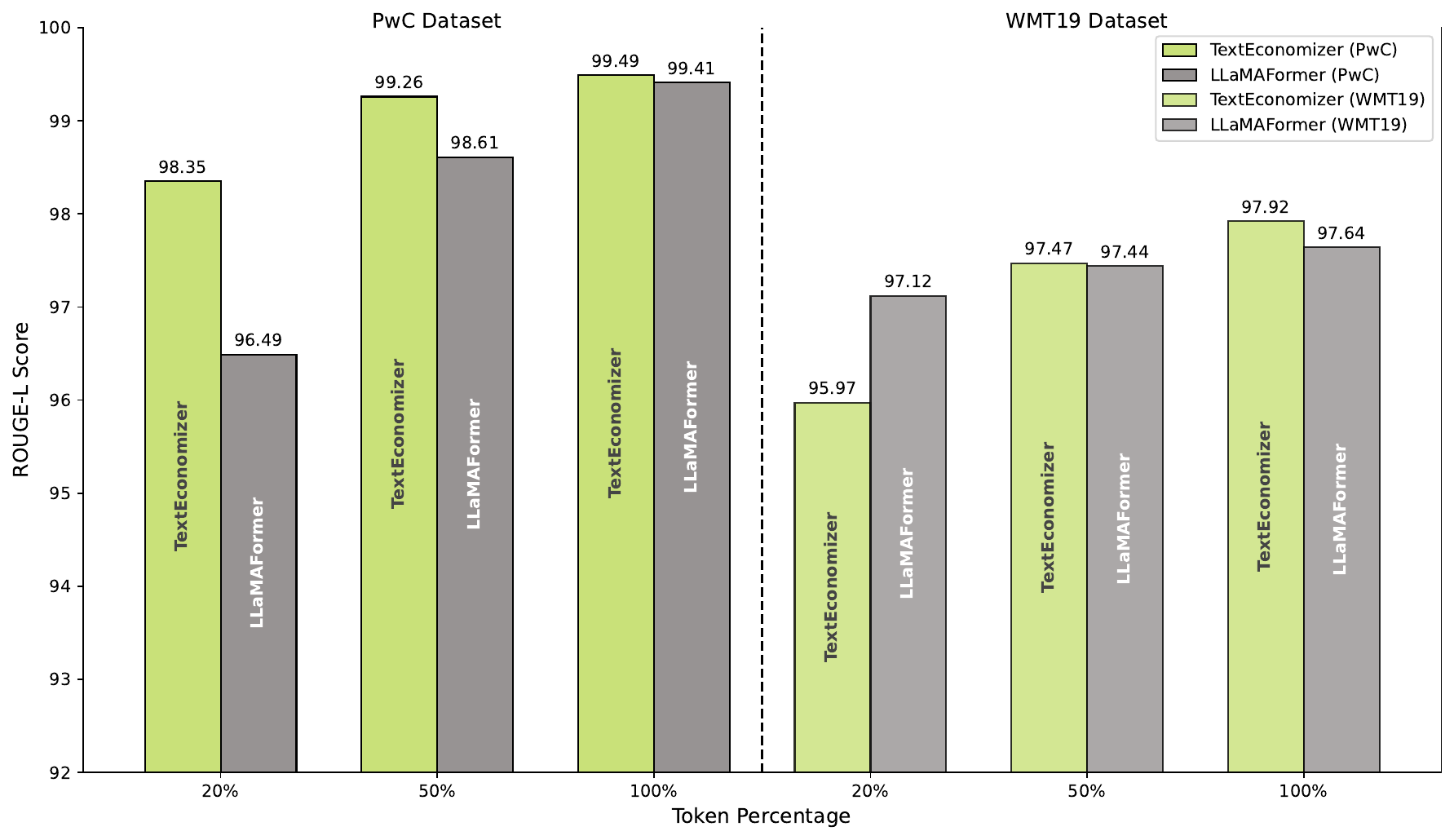}
    \caption{ROUGE-L metrics, elucidating the adeptness of {\methodname} and LLaMAFormer in preserving extended sequence coherence, span token usage gradations (20\%, 50\%, 100\%) on the PwC and WMT19 datasets, with augmented scores echoing superior alignment.}
    \label{fig:ablation_rouge}
\end{figure}

\begin{figure}[t]
    \centering
    \includegraphics[width=0.5\textwidth]{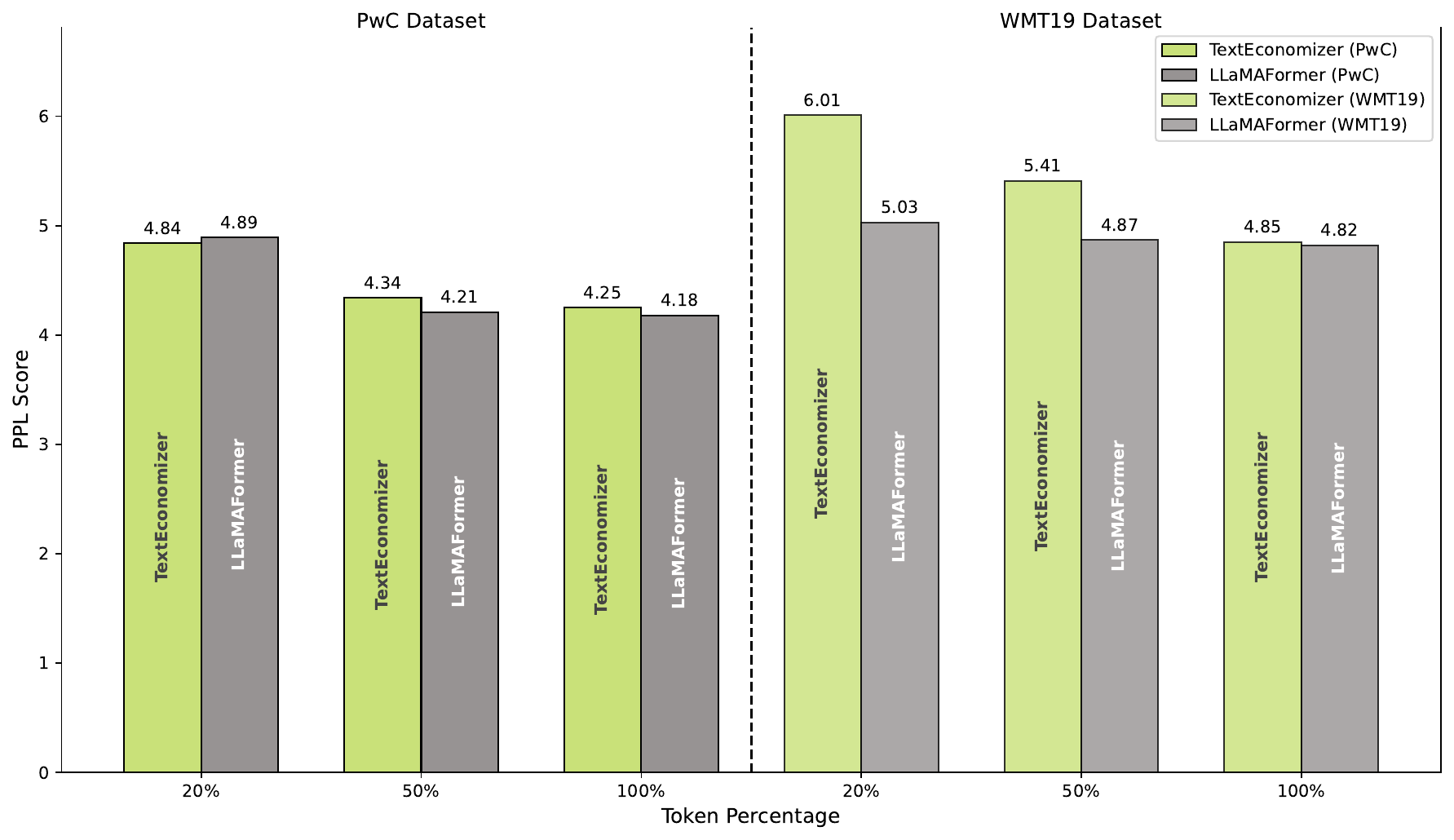}
    \caption{Perplexity trends for {\methodname} and LLaMAFormer across token usage levels (20\%, 50\%, 100\%) on both PwC and WMT19 datasets, with lower scores indicating enhanced language modeling.}
    \label{fig:ablation_ppl}
\end{figure}

\subsubsection*{METEOR \citep{banerjee2005meteor}}
The METEOR (Metric for Evaluation of Translation with Explicit ORdering) metric evaluates the quality of a candidate translation by aligning unigrams with reference translations using exact, stem, and synonym matches. It combines precision and recall into a recall-weighted F-measure and applies a fragmentation penalty to discourage disordered alignments. By emphasizing recall and incorporating stems and synonyms, METEOR captures semantic equivalence more effectively than basic n-gram precision. Scores range from 0 to 100, with higher scores indicating a better match. The mathematical formula for METEOR is as follows:

\begin{equation}
\text{METEOR} 
= \left(1 - \gamma \left(\frac{ch}{m}\right)^{\beta}\right)
  \times 
  \frac{10 \, P \, R}{R + 9P}
\label{eq:meteor_single}
\end{equation}

Here, $P$ and $R$ denote unigram precision and recall from the alignment. $m$ is the number of matched unigrams, and $ch$ is the count of contiguous matched chunks. The tunable parameters are set to \( \gamma = 0.5 \) and \( \beta = 3 \).

\subsubsection*{Perplexity \citep{jelinek1976continuous}}
Perplexity is the exponential of cross-entropy loss, reminiscing how uncertain the model is about the test set, computed as \( \text{PPL} = e^{H(p, q)} \), where \( H(p, q) \) is the cross-entropy between true and predicted distributions. Lower PPL indicates confident, fluent predictions; higher PPL suggests uncertainty or poor generalization. It is widely used to evaluate language models for natural sequence generation.
\begin{table*}[t]
\caption{Comparative results for {\methodname} and LLaMAFormer on PwC and WMT19 using 20\% and 50\% Kizuki vectors. We report BLEU, BERTScore, ROUGE, METEOR, and Perplexity to analyze the effect of Temperature Scaling. Changes against the non-scaled baseline are color-coded: \textcolor{darkgreen}{$\uparrow$} (improvement) and \textcolor{red}{$\downarrow$} (degradation) for semantic quality, with \textcolor{blue}{$\downarrow$} distinguishing fluency-related improvements in Perplexity.}
  \centering
  \begin{adjustbox}{width=\textwidth}
    \begin{tabular}{l||c||c||ccccccc|ccccccc}
      \toprule
      \multirow{3}{*}{\textbf{Models}} 
      & \multirow{3}{*}{\textbf{\#Param}}
      & \multirow{3}{*}{\textbf{Token\%}}
      & \multicolumn{14}{c}{\textbf{PwC}} \\
      \cmidrule(lr){4-17}

      & & 
      & \multicolumn{7}{c|}{\textbf{w/o Temperature Scaling}} 
      & \multicolumn{7}{c}{\textbf{with Temperature Scaling}} \\
      \cmidrule(lr){4-10} \cmidrule(lr){11-17}

      & & 
      & BLEU & BERT & R1 & R2 & RL & METEOR & PPL
      & BLEU & BERT & R1 & R2 & RL & METEOR & PPL \\
      \midrule

      \texttt{TextEconomizer} & 86M & 20\%
      & 94.28 & 98.81 & 97.49 & 94.94 & 97.41 & 93.50 & 5.37
      & 95.61{\color{darkgreen}{($\uparrow$1.33)}} 
      & 99.11{\color{darkgreen}{($\uparrow$0.30)}} 
      & 98.38{\color{darkgreen}{($\uparrow$0.89)}} 
      & 96.62{\color{darkgreen}{($\uparrow$1.68)}} 
      & 98.35{\color{darkgreen}{($\uparrow$0.94)}} 
      & 94.63{\color{darkgreen}{($\uparrow$1.13)}} 
      & 4.85{\color{blue}{($\downarrow$0.52)}} \\

      \texttt{LLaMAFormer} & 50M & 20\%
      & 90.79 & 98.15 & 95.50 & 90.82 & 95.20 & 91.24 & 5.53
      & 92.67{\color{darkgreen}{($\uparrow$1.88)}}
      & 98.56{\color{darkgreen}{($\uparrow$0.41)}}
      & 96.67{\color{darkgreen}{($\uparrow$1.17)}}
      & 93.16{\color{darkgreen}{($\uparrow$2.34)}}
      & 96.50{\color{darkgreen}{($\uparrow$1.30)}}
      & 91.43{\color{darkgreen}{($\uparrow$0.19)}}
      & 4.89{\color{blue}{($\downarrow$0.63)}} \\

      \midrule

      \texttt{TextEconomizer} & 86M & 50\%
      & 97.20 & 99.44 & 99.39 & 98.55 & 99.39 & 95.38 & 4.37
      & 96.98{\color{red}{($\downarrow$0.22)}}
      & 99.41{\color{red}{($\downarrow$0.03)}}
      & 99.21{\color{red}{($\downarrow$0.18)}}
      & 98.33{\color{red}{($\downarrow$0.22)}}
      & 99.26{\color{red}{($\downarrow$0.13)}}
      & 95.24{\color{red}{($\downarrow$0.14)}}
      & 4.34{\color{blue}{($\downarrow$0.03)}} \\

      \texttt{LLaMAFormer} & 50M & 50\%
      & 96.25 & 99.28 & 98.84 & 97.48 & 98.84 & 94.78 & 4.38
      & 95.94{\color{red}{($\downarrow$0.31)}}
      & 99.22{\color{red}{($\downarrow$0.06)}}
      & 98.61{\color{red}{($\downarrow$0.23)}}
      & 97.05{\color{red}{($\downarrow$0.43)}}
      & 98.61{\color{red}{($\downarrow$0.23)}}
      & 95.43{\color{darkgreen}{($\uparrow$0.65)}}
      & 4.21{\color{blue}{($\downarrow$0.17)}} \\

      \midrule\midrule


      \multirow{3}{*}{\textbf{Models}} 
      & \multirow{3}{*}{\textbf{\#Param}}
      & \multirow{3}{*}{\textbf{Token\%}}
      & \multicolumn{14}{c}{\textbf{WMT19}} \\
      \cmidrule(lr){4-17}

      & & 
      & \multicolumn{7}{c|}{\textbf{w/o Temperature Scaling}} 
      & \multicolumn{7}{c}{\textbf{with Temperature Scaling}} \\
      \cmidrule(lr){4-10} \cmidrule(lr){11-17}

      & &
      & BLEU & BERT & R1 & R2 & RL & METEOR & PPL
      & BLEU & BERT & R1 & R2 & RL & METEOR & PPL \\
      \midrule

      \texttt{TextEconomizer} & 86M & 20\%
      & 90.29 & 98.05 & 95.47 & 91.53 & 95.40 & 92.71 & 6.53
      & 90.97{\color{darkgreen}{($\uparrow$0.68)}}
      & 98.22{\color{darkgreen}{($\uparrow$0.17)}}
      & 96.02{\color{darkgreen}{($\uparrow$0.55)}}
      & 92.46{\color{darkgreen}{($\uparrow$0.93)}}
      & 95.97{\color{darkgreen}{($\uparrow$0.57)}}
      & 93.13{\color{darkgreen}{($\uparrow$0.42)}}
      & 6.01{\color{blue}{($\downarrow$0.52)}} \\

      \texttt{LLaMAFormer} & 50M & 20\%
      & 87.30 & 97.48 & 93.70 & 87.89 & 93.49 & 90.84 & 7.39
      & 93.13{\color{darkgreen}{($\uparrow$5.83)}}
      & 98.68{\color{darkgreen}{($\uparrow$1.20)}}
      & 97.14{\color{darkgreen}{($\uparrow$3.44)}}
      & 94.29{\color{darkgreen}{($\uparrow$6.40)}}
      & 97.12{\color{darkgreen}{($\uparrow$3.63)}}
      & 94.85{\color{darkgreen}{($\uparrow$4.01)}}
      & 5.03{\color{blue}{($\downarrow$2.37)}} \\

      \midrule

      \texttt{TextEconomizer} & 86M & 50\%
      & 91.68 & 98.07 & 96.67 & 94.22 & 94.67 & 94.74 & 5.60
      & 93.18{\color{darkgreen}{($\uparrow$1.50)}}
      & 98.67{\color{darkgreen}{($\uparrow$0.60)}}
      & 97.47{\color{darkgreen}{($\uparrow$0.80)}}
      & 95.12{\color{darkgreen}{($\uparrow$0.90)}}
      & 97.47{\color{darkgreen}{($\uparrow$2.80)}}
      & 95.64{\color{darkgreen}{($\uparrow$0.90)}}
      & 5.33{\color{blue}{($\downarrow$0.27)}} \\

      \texttt{LLaMAFormer} & 50M & 50\%
      & 92.30 & 98.56 & 96.86 & 93.91 & 96.85 & 94.15 & 5.49
      & 93.61{\color{darkgreen}{($\uparrow$1.31)}}
      & 98.78{\color{darkgreen}{($\uparrow$0.22)}}
      & 97.45{\color{darkgreen}{($\uparrow$0.59)}}
      & 94.87{\color{darkgreen}{($\uparrow$0.96)}}
      & 97.44{\color{darkgreen}{($\uparrow$0.59)}}
      & 95.46{\color{darkgreen}{($\uparrow$1.31)}}
      & 4.87{\color{blue}{($\downarrow$0.62)}} \\

      \bottomrule
    \end{tabular}
  \end{adjustbox}
  \label{tab:temp_scale_ablation}
\end{table*}
\subsection{Hardware Platform}
The experimental models were implemented using Python 3.10 and trained on Kaggle’s system configuration, which consists of an Intel(R) Xeon(R) CPU @ 2.20GHz (x86\_64 architecture) with 4 vCPU cores \cite{CONSOLVO2023Hardware}. To accelerate training, we utilized a single NVIDIA Tesla P100 GPU with 16 GB of HBM2 VRAM. For data processing and model handling, we leveraged the Kaggle notebook’s upgraded 30 GB of system RAM, ensuring sufficient memory for the experimental pipeline. The models were developed using PyTorch 2 and associated torch libraries.

\subsection{Ablation Study}

In this thorough ablation study, we investigate three pivotal dimensions of model efficiency—corpus size, leveraging {\vectorname} vectors, and integrating traditional and trendy transformer architectural components across our proposed framework, {\methodname}. Additionally, we explore LLaMAFormer, a lightweight variant of the transformer-based encoder-decoder method that incorporates essential architectural elements from LLaMA. In this model, we applied RMSNorm immediately after the embedding layer and before rotary positional encoding of the query $(Q)$ and key $(K)$ vectors, effectively replacing absolute positional encoding. Residual connections were introduced after the multi-head attention layers to enhance training stability, and the feed-forward network now employs SwiGLU instead of the conventional ReLU activation. With all traditional Layer Normalizations replaced by RMSNorm for improved computational efficiency, LLaMAFormer retains the core flow of {\methodname}. Subsequently, as a baseline, we adopt an Autoencoder model that mirrors the implementation of \cite{bahdanau2014neural}, substituting the original bidirectional RNN \cite{schuster1997bidirectional} with a bidirectional Gated Recurrent Unit (Bi-GRU) \cite{chung2014empirical}, which is an integral component of our framework, and evaluating exclusively on English autoencoding tasks. 

Another key focus of our study is to assess the role of attention in sequence-to-sequence (seq2seq) learning by comparing performance metrics with and without the attention mechanism in the Autoencoder. Empirical results on the WMT19 and PwC datasets demonstrate that {\methodname} invariably outperforms both the Autoencoder baseline and LLaMAFormer across several evaluation metrics, highlighting the consequence of our architectural preferences and providing insights into optimizing efficiency in modern neural autoencoding systems.
First, as illustrated in Fig. [\ref{fig:ablation_bleu}, \ref{fig:ablation_bert}, \ref{fig:ablation_rouge}], we evaluated the efficacy of token reduction by conducting experiments with three configurations—20\%, 50\%, and 100\% {\vectorname} vectors usage across two datasets (PwC and WMT19). The results indicate that {\methodname} achieves highly competitive scores even with substantial token reductions, highlighting its ability to focus on the most relevant portions of the encoder output. For instance, at 20\% {\vectorname} vector usage, {\methodname} achieves a BLEU score of 95.61 on the PwC dataset, which is only marginally inferior to its full-token counterpart. Similarly, on the WMT19 dataset, the 20\% configuration yields a BLEU score of 90.97, maintaining potent performance despite passing only a fraction of the {\vectorname}s to the decoder. These findings are further corroborated by the BERTScore, METEOR, and ROUGE-L metrics, where {\methodname} consistently outperforms or closely matches the performance of LLaMAFormer under similar token constraints. Moreover, when augmenting {\vectorname} vector usage to 50\%, {\methodname} exhibits near-optimal performance, achieving a BLEU score of 96.98 on the PwC dataset and 93.19 on WMT19. Notably, these scores are within 0.35 and 1.04 points of the full-token configuration, respectively, while significantly reducing computational overhead. This suggests that our framework effectively prioritizes the most salient information from the encoder output, thereby enhancing the efficiency of the cross-attention mechanism without compromising quality.

The perplexity (PPL) results, portrayed in Fig. \ref{fig:ablation_ppl}, further validate the efficacy of {\vectorname} vectors. On the PwC dataset, {\methodname} achieves a PPL of 4.34 at 50\% {\vectorname} vector usage, closely identical to the full-token PPL of 4.25. Likewise, on WMT19, the 50\% configuration produces a PPL of 5.33, closely matching the full-token PPL of 4.85. These observations underscore the robustness of {\methodname} in leveraging reduced context vectors to maintain high-quality predictions. In addition to these observations, LLaMAFormer exhibits distinctive advantages in high-sparsity regimes on the WMT19 dataset. At 20\% {\vectorname} vectors, it acquires a BLEU score of 93.13, outperforming {\methodname}’s 90.97 under the same settings while utilizing 1.72$\times$ fewer parameters. In contrast, its full-token PPL (4.82) marginally exceeds {\methodname}’s 4.85 on WMT19, suggesting stronger fluency in certain autoencoding configurations. However, {\methodname} dominates in token efficiency on PwC, with a \texttt{+2.94} BLEU gain over LLaMAFormer at 20\% {\vectorname} vectors.
\begin{table}[t]
\caption{Comparison of compression ratio, decoding speed, and computational cost across different token percentages for NUGGET, TextEconomizer, LLaMAFormer, and AutoEncoder. $\Theta$ denotes no compression.}
\centering
\begin{adjustbox}{width=0.48\textwidth}
\begin{tabular}{l||c||c|c|c|c|c}
\toprule
\textbf{Methods} & \textbf{\#Params} & \textbf{Token\%} & \textbf{Compression Ratio} & \textbf{Decoding Speed (KB/s)} & \textbf{GFLOPs} & \textbf{MACs} \\
\midrule

\texttt{NUGGET}
  & 602M
  & 10  & 10$\times$ & 38  & 85.48 & 42.74$\times 10^{9}$ \\
\midrule

\multirow{3}{*}{\texttt{TextEconomizer}}
  & \multirow{3}{*}{86M} 
  & 20  & 5.39$\times$ & 146  & 7.22 & 3.61$\times 10^{9}$ \\
  &    & 50  & 3.86$\times$ & 121  & 7.35 & 3.68$\times 10^{9}$ \\
  &    & 100 & $\Theta$    & 75   & 7.58 & 3.79$\times 10^{9}$ \\
\midrule

\multirow{3}{*}{\texttt{LLaMAFormer}}
  & \multirow{3}{*}{50M}
  & 20  & 5.39$\times$ & 133  & 8.83 & 4.42$\times 10^{9}$ \\
  &    & 50  & 3.86$\times$ & 107  & 8.97 & 4.48$\times 10^{9}$ \\
  &    & 100 & $\Theta$    & 60   & 9.20 & 4.60$\times 10^{9}$ \\
\midrule

\texttt{AutoEncoder}
  & 67M 
  & 100 & 67$\times$ & 5 & 13.27 & 6.64$\times 10^{9}$ \\
\bottomrule
\end{tabular}
\end{adjustbox}
\label{tab:compression_speed_compute}
\end{table}

Afterward, our experimental results (Table \ref{tab:temp_scale_ablation}) indicate that temperature scaling is a key calibration technique that enhances fluency and consistently lowers Perplexity (PPL) scores across all configurations, effectively refining the model's output distribution. Our initial finding is that temperature scaling enhances semantic fidelity, particularly in high-sparsity scenarios (20\% token retention). Scaling stabilizes against context loss, leading to consistent gains across all the metrics and datasets. For example, LLaMAFormer on the WMT19 benchmark shows a notable improvement of \texttt{+5.83} BLEU, indicating that scaling effectively recalibrates erratic raw logits and prevents significant errors. In lower-sparsity settings with 50\% token retention, a positive trend is evident, especially with the WMT19 dataset, which continues to improve. Most performance metrics reflect this, though the PwC dataset shows an insignificant saturation point without further scaling. In this context, minor drops occur in performance, such as a \texttt{-0.22} BLEU decrease for {\methodname}. However, many other semantic metrics remain stable or improve; for instance, LLaMAFormer’s BLEU score declines, but its METEOR score increases by \texttt{+0.65}, indicating better semantic alignment. Overall, despite some minor and localized decreases, temperature scaling demonstrates its effectiveness as a crucial component in robust token selection.

Moreover, our analysis of computational efficiency, summarized in Table \ref{tab:compression_speed_compute}, reveals that TextEconomizer consistently offers faster decoding and requires fewer floating point operations (FLOPs) and multiply-accumulate operations (MACs) across all token ratios. For example, at 20\% token retention, TextEconomizer achieves 146 KB/s decoding speed compared to 133 KB/s for LLaMAFormer, while using 7.22 GFLOPs versus 8.83 GFLOPs. Notably, TextEconomizer outperforms LLaMAFormer, decoding at 75 KB/s with 7.58 GFLOPs compared to LLaMAFormer's 60 KB/s and 9.20 GFLOPs, even at 100\% token percentage. While NUGGET, operating at 10\% token retention, achieves a strong 10$\times$ compression ratio, it incurs substantially higher computational cost at 85.48 GFLOPs and 42.74$\times 10^{9}$ MACs, with a decoding speed of 38 KB/s. This exposes a key architectural bottleneck in large-scale encoder-decoder models: because feed-forward network layers account for the overwhelming majority of total computation and are independent of the number of retained tokens, aggressive token compression translates to negligible reductions in actual FLOPs. In contrast, the AutoEncoder achieves a higher compression ratio (67$\times$) but incurs substantially slower decoding (4.42 KB/s) and higher computational cost (13.27 GFLOPs). Overall, these results emphasize the efficiency of token-based compression over latent-space autoencoding methods when semantic fidelity is the key factor.
\renewcommand{\arraystretch}{1.4}
\begin{table}[t]
\caption{The influence of corpus size on the performance of the proposed AutoEncoder.}
\centering
\begin{adjustbox}{width=0.48\textwidth}
\begin{tabular}{lccccc}
\hline
\multirow{2}{*}{\textbf{Method}} & \multirow{2}{*}{\textbf{Corpus Size}} & \multicolumn{4}{c}{\textbf{Inference}} \\
\cline{3-6}
 & & \textbf{BLEU} & \textbf{BERT Score} & \textbf{R-L} & \textbf{PPL} \\
\hline
\multirow{3}{*}{\texttt{AutoEncoder}}
  & 100K       & 87.30 & 97.27 & 93.69 & 22.44 \\
  & 200K       & 91.97 & 98.41 & 96.59 & 17.92 \\
  & 242K       & 95.75 & 99.27 & 98.85 & 11.26 \\
\hline
\multirow{2}{*}{\texttt{AutoEncoder (No Attention)}}
  & 242K (PwC)   & 7.87  & 79.78 & 19.29 & 948.85 \\
  & 600K (WMT19) & 31.16 & 85.68 & 46.72 & 240.02 \\
\hline
\end{tabular}
\end{adjustbox}
\label{tab:ablation_study}
\end{table}
Additionally, we isolate LZMA's contribution from the architectural gains across all datasets. In the Autoencoder, the neural component has a significant impact: PwC achieves a total compression of 67$\times$ (with the Autoencoder contributing 64.33$\times$ and LZMA contributing 2.67$\times$). 
\begin{table}[!ht]
\caption{Performance of TextEconomizer under noisy test conditions on the PwC and WMT19 datasets at 50\% and 100\% Kizuki vector retention. Values in parentheses denote marginal changes relative to the clean test baseline. Red arrows indicate performance degradation: {\color{red}$\downarrow$} for quality metrics (BLEU, BERTScore, ROUGE, METEOR) and {\color{red}$\uparrow$} for perplexity (PPL, where lower is better).}
  \centering
  \begin{adjustbox}{width=0.50\textwidth}
    \begin{tabular}{l c c c c c c c}
      \toprule
      \multirow{2}{*}{\textbf{Models}} & \multicolumn{7}{c}{\textbf{PwC}} \\
      \cmidrule(lr){2-8}
       & BLEU & BERTScore & ROUGE-1 & ROUGE-2 & ROUGE-L & METEOR & PPL \\
      \midrule
      \verb|TextEconomizer 50%|  & 95.41 {\color{red}(1.57) $\downarrow$} & 99.18 {\color{red}(0.23) $\downarrow$} & 98.73 {\color{red}(0.48) $\downarrow$} & 97.61 {\color{red}(0.72) $\downarrow$} & 98.79 {\color{red}(0.47) $\downarrow$} & 94.58 {\color{red}(0.66) $\downarrow$} & 4.52 {\color{red}(0.18) $\uparrow$} \\
      \verb|TextEconomizer 100%| & 95.87 {\color{red}(1.46) $\downarrow$} & 99.21 {\color{red}(0.25) $\downarrow$} & 98.93 {\color{red}(0.56) $\downarrow$} & 97.96 {\color{red}(0.82) $\downarrow$} & 98.91 {\color{red}(0.58) $\downarrow$} & 98.19 {\color{red}(0.68) $\downarrow$} & 4.40 {\color{red}(0.15) $\uparrow$} \\
      \midrule[\heavyrulewidth]

      \multirow{2}{*}{\textbf{Models}} & \multicolumn{7}{c}{\textbf{WMT19}} \\
      \cmidrule(lr){2-8}
       & BLEU & BERTScore & ROUGE-1 & ROUGE-2 & ROUGE-L & METEOR & PPL \\
      \midrule
      \verb|TextEconomizer 50%|  & 91.74 {\color{red}(1.44) $\downarrow$} & 98.41 {\color{red}(0.26) $\downarrow$} & 96.89 {\color{red}(0.58) $\downarrow$} & 94.31 {\color{red}(0.81) $\downarrow$} & 96.83 {\color{red}(0.64) $\downarrow$} & 94.93 {\color{red}(0.71) $\downarrow$} & 5.57 {\color{red}(0.24) $\uparrow$} \\
      \verb|TextEconomizer 100%| & 92.57 {\color{red}(1.56) $\downarrow$} & 98.58 {\color{red}(0.28) $\downarrow$} & 97.27 {\color{red}(0.65) $\downarrow$} & 94.96 {\color{red}(0.88) $\downarrow$} & 97.29 {\color{red}(0.63) $\downarrow$} & 94.17 {\color{red}(0.72) $\downarrow$} & 5.16 {\color{red}(0.31) $\uparrow$} \\

      \bottomrule
    \end{tabular}
  \end{adjustbox}
  \label{tab:noisy_test}
\end{table}
For WMT19 and WMT14, the total reaches approximately 33$\times$ (where the Autoencoder accounts for about 31.7$\times$ and LZMA contributes between 1.25$\times$ and 1.34$\times$). BookCorpus achieves a total of 25$\times$ (with the Autoencoder contributing 23.14$\times$ and LZMA contributing 1.86$\times$). 
\renewcommand{\arraystretch}{1.4}
\begin{table*}[t]
\caption{The comparison of the quantitative performance of various existing methods across PwC and WMT19 datasets. In this table, $\mathbf{r}$ denotes the memory compression ratio, while $\mathbf{\Theta}$ symbolizes no memory savings.}
    \centering
    \begin{adjustbox}{width=0.95\textwidth}
    \begin{tabular}{|l|c|c||cccccccc||cccccccc||}
        \hline
        \multirow{2}{*}{\textbf{Method}} & \multirow{2}{*}{\textbf{\#Params.}} & \multirow{2}{*}{\textbf{Token\%}} &
        \multicolumn{8}{c||}{PwC} & \multicolumn{8}{c||}{WMT19} \\
        \cline{4-19}
         {} & {} & {} &
         \textbf{BLEU} & \textbf{BERT Score} & \textbf{R1} & \textbf{R2} & \textbf{R-L} & \textbf{METEOR} & \textbf{PPL} & $\mathbf{r}$ &
         \textbf{BLEU} & \textbf{BERT Score} & \textbf{R1} & \textbf{R2} & \textbf{R-L} & \textbf{METEOR} & \textbf{PPL} & $\mathbf{r}$ \\
        \hline

        \texttt{ICAE}  & 13.13B & 100\% &
        \textbf{99.8} & $-$ & $-$ & $-$ & $-$ & $-$ & 9.50 & 4$\times$ &
        $-$ & $-$ & $-$ & $-$ & $-$ & $-$ & $-$ & $-$ \\

        \texttt{NUGGET} & {602M} & 10\% &
        $-$ & $-$ & $-$ & $-$ & $-$ & $-$ & $-$ & $-$ &
        \textbf{99} & $-$ & $-$ & $-$ & $-$ & $-$ & 28.10 & 10$\times$ \\

        \texttt{T5-Small} & {70M} & 100\% &
        38.29 & 93.58 & 66.92 & 65.12 & 66.89 & 62.20 & 1.04 & $\Theta$ &
        50.15 & 94.55 & 75.99 & 73.40 & 75.95 & 68.36 & 1.07 & $\Theta$ \\
        \hline

        \multicolumn{19}{|c||}{\textbf{Our Proposed Methods}} \\
        \hline

        \texttt{{\methodname}} & {86M} & 100\% &
        97.33 & 99.46 & 99.49 & 98.78 & 99.49 & 98.87 & 4.25 & 1.32$\times$ &
        94.13 & 98.86 & 97.92 & 95.84 & 97.92 & 94.89 & 4.85 & 1.14$\times$ \\

        \texttt{Autoencoder} & 67M & 100\% &
        95.75 & 99.28 & 98.86 & 97.66 & 98.85 & 96.31 & 11.26 & \textbf{67$\times$} &
        91.94 & 98.41 & 96.65 & 93.73 & 96.61 & 92.75 & 22.25 & \textbf{33$\times$} \\

        \texttt{LLaMAFormer} & \textbf{50M} & 50\% &
        95.94 & 99.22 & 98.61 & 97.05 & 98.61 & 95.43 & 4.21 & 3.86$\times$ &
        93.61 & 98.78 & 97.45 & 94.87 & 97.44 & 95.46 & 4.87 & 3.08$\times$ \\

        \texttt{{\methodname}} & {86M} & 50\% &
        96.98 & 99.41 & 99.27 & 98.33 & 99.26 & 95.24 & 4.34 & 3.86$\times$ &
        93.18 & 98.67 & 97.47 & 95.12 & 97.47 & 95.64 & 5.33 & 3.08$\times$ \\
        \hline
    \end{tabular}
    \end{adjustbox}
    \label{tab:quantitative_res_1}
\end{table*}


\renewcommand{\arraystretch}{1.4}
\begin{table*}[t]
\caption{The comparison of the quantitative performance of various existing methods across WMT14 and BookCorpus datasets. In this table, $\mathbf{r}$ denotes the memory compression ratio, while $\mathbf{\Theta}$ symbolizes no memory savings.}
    \centering
    \begin{adjustbox}{width=0.95\textwidth}
    \begin{tabular}{|l|c|c||cccccccc||cccccccc||}
        \hline
        \multirow{2}{*}{\textbf{Method}} &
        \multirow{2}{*}{\textbf{\#Params.}} &
        \multirow{2}{*}{\textbf{Token\%}} &
        \multicolumn{8}{c||}{WMT14} &
        \multicolumn{8}{c||}{BookCorpus} \\
        \cline{4-19}
        {} & {} & {} &
        \textbf{BLEU} & \textbf{BERT Score} & \textbf{R1} & \textbf{R2} & \textbf{R-L} & \textbf{METEOR} & \textbf{PPL} & $\mathbf{r}$ &
        \textbf{BLEU} & \textbf{BERT Score} & \textbf{R1} & \textbf{R2} & \textbf{R-L} & \textbf{METEOR} & \textbf{PPL} & $\mathbf{r}$ \\
        \hline

        \texttt{T5-Small} & {70M} & 100\% &
        60.05 & 95.45 & 81.52 & 79.31 & 81.52 & 70.92 & 1.16 & $\Theta$ &
        73.37 & 97.04 & 92.16 & 88.47 & 92.15 & 75.28 & 1.03 & $\Theta$ \\
        \hline

        \multicolumn{19}{|c||}{\textbf{Our Proposed Method}} \\
        \hline

        \texttt{{\methodname}} & {86M} & 100\% &
        94.08 & 98.93 & 97.84 & 95.51 & 97.84 & 94.76 & 4.90 & 1.25$\times$ &
        90.58 & 97.92 & 93.89 & 87.11 & 93.86 & 91.32 & 5.24 & 1.12$\times$ \\

        \texttt{Autoencoder} & 67M & 100\% &
        91.72 & 98.39 & 96.39 & 93.37 & 96.37 & 92.39 & 21.52 & \textbf{33$\times$} &
        88.81 & 97.61 & 92.90 & 85.15 & 92.86 & 90.12 & 12.33 & \textbf{25$\times$} \\

        \texttt{LLaMAFormer} & \textbf{50M} & 50\% &
        93.60 & 98.74 & 97.51 & 94.94 & 97.50 & 94.15 & 5.04 & 3.07$\times$ &
        89.92 & 97.82 & 93.52 & 86.37 & 93.48 & 90.96 & 5.31 & 3.10$\times$ \\

        \texttt{{\methodname}} & {86M} & 50\% &
        94.18 & 99.06 & 98.25 & 96.42 & 98.25 & 94.85 & 4.95 & 3.07$\times$ &
        90.30 & 97.88 & 93.71 & 86.75 & 93.66 & 91.14 & 5.33 & 3.10$\times$ \\
        \hline
    \end{tabular}
    \end{adjustbox}
    \label{tab:quantitative_res_2}
\end{table*}

For TextEconomizer/LLaMAFormer, token pruning sets the baseline. At 20\% token retention (4$\times$ the baseline), LZMA adds 1.39$\times$ for a total of 5.39$\times$. At 50\% retention (2$\times$ the baseline), LZMA contributes 1.86$\times$ for PwC and between 1.07$\times$ and 1.10$\times$ for other datasets. At 100\% retention (using LZMA only), the ratios are modest, ranging from 1.12$\times$ to 1.32$\times$. We attribute this modest performance to the high entropy of continuous values of the tensors compared to structured text. Nevertheless, we retain LZMA since these incremental gains can add up significantly in high-frequency decoding scenarios. Furthermore, as depicted in Table \ref{tab:ablation_study}, scaling the PwC corpus size from 100K to 242K instances in the Autoencoder neural network significantly enhances performance across all metrics. The full 242K corpus achieves a BLEU score of 95.75 and a PPL of 11.26, outperforming smaller corpora. This trend aligns with findings from \citep{bijoy-etal-2023-advancing}, confirming that larger training data volumes improve linguistic coherence and task-specific accuracy. To rigorously evaluate architectural distinctions, we compared the Autoencoder to our {\methodname} with an equivalent parameter count. The transformer achieved superior results, including a BLEU score of 97.43, a BERT Score of 99.48, and a ROUGE-L score of 99.54, accentuating its advantages over sequential architectures like LSTMs. Notably, ablating the attention mechanism which is a core component of both architectures, harshly decreases performance (e.g., PwC BLEU drops to 7.87), underscoring its paramount role in contextual alignment. While the Autoencoder model lags behind transformer-based architectures like {\methodname} and LLaMAFormer in task-specific metrics, its fixed-size bottleneck layer offers a special advantage in memory efficiency. By compressing inputs into a fixed-size latent space, the Autoencoder reduces computational overhead greatly, making it a practical choice for scenarios prioritizing resource conservation over state-of-the-art semantic performance. This trade-off emphasizes the importance of architectural flexibility: transformer-based models excel in quality-driven applications, whereas the Autoencoder provides a lightweight alternative for environments with stringent memory limitations.

Synthesizing these insights, {\methodname} emerges as a pragmatic choice for resource-constrained deployments, with a minimal 1.72 BLEU drop on PwC at 50\% {\vectorname} vectors. Likewise, LLaMAFormer exhibits robustness in high-sparsity regimes, particularly on WMT19. Both models validate token reduction as a scalable strategy for English autoencoding tasks, provided attention mechanisms remain consistent, hovering computational efficiency with output quality. The Autoencoder, while lower in metrics (e.g., 95.75 BLEU vs. 97.33), offers better memory efficiency and is suitable for resource-conserving deployments, emphasizing the choice between quality and footprint.

\renewcommand{\arraystretch}{1.6}
\begin{table*}[ht!]
\small
\caption{The qualitative effectiveness of various transformer-based methods in contrast to TextEconomizer,
with per-output \textbf{B}~(BLEU), \textbf{RL}~(ROUGE-L), \textbf{M}~(METEOR), and \textbf{BS}~(BERTScore) computed against the (Input) reference.
\colorbox{red!30}{\textbf{Red}} denotes ignored/wrong/extra words/characters.
``--'' indicates the reference row (scores are trivially 1).}
\centering
\begin{adjustbox}{width=\textwidth}
\begin{tabular}{l p{1.10\textwidth} r r r r}
\hline
\textbf{} & \textbf{Text} & \textbf{B} & \textbf{RL} & \textbf{M} & \textbf{BS} \\
\hline

(Input) &
{reid and partner alfie hewett came from a set down to beat the french pair stephane houdet and nicolas peifer 4-66-1 7-6(8-6).}
& -- & -- & -- & -- \\

(ICAE) &
{reid and \colorbox{red!30}{partner} alfie hewett came from a set down to beat the french pair stephane houdet and nicolas peifer 4-66-1 7-6(8-6).}
& 1.00 & 1.00 & 1.00 & 1.00 \\

(T5 Small) &
{reid and partner alfie hewett came from a set down to beat the french pair \colorbox{red!30}{stephan}}
& 0.30 & 0.60 & 0.47 & 0.83 \\

(Autoencoder) &
{reid and partner alfie hewett came from a set down to beat the french pair stephane houdet and nicolas peifer 4-66\colorbox{red!30}{-6-8}(8-6)}
& 0.82 & 0.91 & 0.92 & 0.95 \\

(LLaMAFormer) &
{reid and partner alfie hewett came from a set down to beat the french pair stephane houdet and nicolas peifer 4-66-1 7-6(8-6).}
& 1.00 & 1.00 & 1.00 & 1.00 \\

({\methodname}) &
{reid and partner alfie hewett came from a set down to beat the french pair stephane houdet and nicolas peifer 4-66-1 7-6(8-6).}
& 1.00 & 1.00 & 1.00 & 1.00 \\

\hline

(Input) &
{experimentally, we comprehensively compare the behavior of icl and explicit fine-tuning based on real tasks to provide empirical evidence that supports our understanding. the results prove that icl behaves similarly to explicit fine-tuning at the prediction level, the representation level, and the attention behavior level.}
& -- & -- & -- & -- \\

(ICAE) &
{experimentally, we comprehensively compare the behavior of icl and explicit finetuning based on real tasks to provide empirical evidence that supports our \colorbox{red!30}{findings}. the \colorbox{red!30}{experimental evidence proves} that icl behaves \colorbox{red!30}{like us to the same extent}. \colorbox{red!30}{prediction} at the explicit \colorbox{red!30}{finetuning} level, the representation level, and the attention behavior level.}
& 0.60 & 0.76 & 0.81 & 0.91 \\

(T5 Small) &
{experimentally, we comprehensively compare the behavior of icl and explicit fine-tuning}
& 0.06 & 0.41 & 0.28 & 0.66 \\

(Autoencoder) &
{experimentally, we comprehensively compare the behavior of icl and explicit fine-tuning based on real tasks to provide empirical evidence that supports our understanding. the results prove that icl behaves similarly to explicit fine-tuning at the prediction level, the representation level, and the attention behavior level.}
& 1.00 & 1.00 & 1.00 & 1.00 \\

(LLaMAFormer) &
{experimentally, we comprehensively compare the behavior of icl and explicit fine-tuning based on real tasks to provide empirical evidence that supports our understanding. the results prove that icl behaves similarly to explicit fine-tuning at the prediction level, the representation level, and the attention behavior level.}
& 1.00 & 1.00 & 1.00 & 1.00 \\

({\methodname}) &
{experimentally, we comprehensively compare the behavior of icl and explicit fine-tuning based on real tasks to provide empirical evidence that supports our understanding. the results prove that icl behaves similarly to explicit fine-tuning at the prediction level, the representation level, and the attention behavior level.}
& 1.00 & 1.00 & 1.00 & 1.00 \\

\hline

(Input) &
{i was eagerly preparing for my trip to paris, france, where the eiffel tower stands tall, to attend a conference on machine learning, a subset of artificial intelligence, scheduled from september 10 to september 12, 2023. i have not read that book yet-the one recommended for the conference-but i plan to do so soon, perhaps during the flight. back home, despite the heavy rain, which had been pouring since morning, the football match continued as scheduled, much to the delight of the fans, and i couldn't help but follow the updates online. the company that sponsors the team had just announced that its revenue increased by 15\% last quarter, reaching \$10 million, a significant milestone, thanks to their cutting-edge ai innovations-some of which would be showcased at the conference. on the plane to paris, i overheard a fellow passenger ask, `what is the capital of australia, and why is it not sydney?'-prompting me to recall that it's canberra, designed specifically as the capital, unlike the more famous sydney. unfortunately, when i arrived in paris, i felt under the weather and decided to stay in my hotel, missing the first day of the conference. as i rested, a quirky thought crossed my mind: `time flies like an arrow; fruit flies like a banana,' and i smiled, hoping to recover quickly for the remaining days.}
& -- & -- & -- & -- \\

(T5 Small) &
{i was eagerly preparing for my trip to paris, france \colorbox{red!30}{:} where the eiffel tower stands tall, to attend a conference on machine learning, a subset}
& 0.00 & 0.18 & 0.12 & 0.52 \\

(Autoencoder) &
{i was eagerly preparing for my trip to paris, france, where the eiffel tower stands \colorbox{red!30}{tall} to \colorbox{red!30}{to a} a conference on machine learning, a subset of artificial intelligence, scheduled from september 10 to september 12, 2023. i have not read that book yet-the one recommended for the conference-but i plan to do so soon, perhaps during the \colorbox{red!30}{flight} back \colorbox{red!30}{home.} despite the heavy rain, which had been pouring since morning, the football match continued as scheduled, much to the delight of the fans, and i \colorbox{red!30}{couldn} help \colorbox{red!30}{help follow} follow the updates online. the company that sponsors the team had just announced that its revenue increased by 15\% last quarter, reaching \$10 million, a significant milestone, thanks to their cutting-edge ai innovations-some of which would be showcased at the conference. on the plane to paris, i overheard a fellow passenger \colorbox{red!30}{ask`, what} is the capital of australia, and why is it not \colorbox{red!30}{sydney'-prompting} me to recall that \colorbox{red!30}{it} canberra, designed specifically as the capital, unlike the more famous sydney. unfortunately, when i arrived in paris, i felt under the weather and decided to stay in my hotel, missing the first day of the conference. as i rested, a quirky thought crossed my \colorbox{red!30}{mind', time files} like an \colorbox{red!30}{arrow,} fruit flies like a \colorbox{red!30}{banana, and} and i \colorbox{red!30}{am,} hoping to recover quickly for the remaining days.}
& 0.87 & 0.94 & 0.95 & 0.99 \\

(LLaMAFormer) &
{i was eagerly preparing for my trip to paris, france, where the eiffel tower stands tall, to attend a conference on machine learning, a subset of artificial intelligence, scheduled from september 10 to september 12, 2023. i have not read that book yet-the one recommended for the conference-but i plan to do so soon, perhaps during the \colorbox{red!30}{flight} back \colorbox{red!30}{home.} despite the heavy rain, which had been pouring since morning, the football match continued as scheduled, much to the delight of the fans, and i couldn't help but follow the updates online. the company that sponsors the team had just announced that its revenue increased by 15\% last quarter, reaching \$10 million, a significant milestone, thanks to their cutting-edge ai innovations-some of which would be showcased at the conference. on the plane to paris, i overheard a \colorbox{red!30}{quirky} sydney. unfortunately, when i arrived in paris, i felt under the weather and decided to stay in my hotel, missing the first day of the conference. as i rested, a quirky thought crossed my \colorbox{red!30}{mind'}. time flies like an \colorbox{red!30}{arrow} fruit flies like a \colorbox{red!30}{banana',} and \colorbox{red!30}{i smiled quickly for the conference. on the plane to paris, i rested a quirky thought-to} recover quickly for the remaining days.}
& 0.78 & 0.87 & 0.85 & 0.96 \\

({\methodname}) &
{i was eagerly preparing for my trip to paris, france, where the eiffel tower stands tall, to attend a conference on machine learning, a subset of artificial intelligence, scheduled from september 10 to september 12, 2023. i have not read that book yet-the one \colorbox{red!30}{was} recommended for the conference-but i plan to do so soon, perhaps during the \colorbox{red!30}{flight} back \colorbox{red!30}{home.} despite the heavy rain, which had been pouring since morning, the football match continued as scheduled, much to the delight of the fans, and i couldn't help but follow the updates online. the company that sponsors the team had just announced that its revenue increased by 15\% last quarter, reaching \$10 million, a significant milestone, thanks to their cutting-edge ai innovations-some of which would be showcased at the conference. on the plane to paris, i overheard a fellow passenger \colorbox{red!30}{ask`, what} is the capital of australia, and why is it \colorbox{red!30}{was} not \colorbox{red!30}{sydney's-prompting} me to recall that it's canberra, designed specifically as the capital, unlike the more famous sydney. unfortunately, when i arrived in paris, i felt under the weather and decided to stay in my hotel, missing the first day of the conference. as i rested, a quirky thought \colorbox{red!30}{my mind: time} flies like an \colorbox{red!30}{arrow,} fruit flies like a \colorbox{red!30}{banana',} and i smiled, hoping to recover quickly for the remaining days.}
& 0.90 & 0.97 & 0.97 & 0.99 \\

\hline
\end{tabular}
\end{adjustbox}
\label{tab:qualitative_res}
\end{table*}

\renewcommand{\arraystretch}{1.6}
\addtocounter{table}{-1}
\begin{table*}[ht!]
\small
\caption{(Cont'd) The qualitative effectiveness of various transformer-based methods in contrast to TextEconomizer.}
\centering
\begin{adjustbox}{width=\textwidth}
\begin{tabular}{l p{1.10\textwidth} r r r r}
\hline
\textbf{} & \textbf{Text} & \textbf{B} & \textbf{RL} & \textbf{M} & \textbf{BS} \\
\hline

(Input) &
{sarah found a \$50 bill on the street and excitedly shouted, ``i'm going to save this!'' ten minutes later, she walked out of the store with \$75 worth of things she didn't need, proudly calling it an ``investment.''}
& -- & -- & -- & -- \\

(T5 Small) &
{sarah found a \$50 bill on the street and excitedly shouted\colorbox{red!30}{, ``} i'm going}
& 0.14 & 0.50 & 0.36 & 0.64 \\

(Autoencoder) &
{sarah found a \$50 bill on the street and excitedly shouted, \colorbox{red!30}{`}i'm going to save this!\colorbox{red!30}{'} ten minutes later, she walked out of the store with \$75 worth of things she didn't need, proudly calling it an \colorbox{red!30}{investment investment.}}
& 0.83 & 0.93 & 0.91 & 0.94 \\

(LLaMAFormer) &
{sarah found a \$50 bill on the street and excitedly \colorbox{red!30}{shouted``, is} going to save this, minor ten minutes \colorbox{red!30}{later.} she walked out of the store with \$75 worth of things she \colorbox{red!30}{didns t} need, proudly calling it an \colorbox{red!30}{``investment investment.}}
& 0.69 & 0.85 & 0.90 & 0.90 \\

({\methodname}) &
{sarah found a \$50 bill on the street and excitedly shouted, ``i'm going to save this\colorbox{red!30}{?}'' ten minutes later, she walked out of the store with \$75 worth of things she didn't need, proudly calling it an ``investment.''}
& 0.95 & 0.98 & 0.98 & 0.99 \\

\hline

(Input) &
{tiny toes and button nose, a bundle of joy soon to expose!}
& -- & -- & -- & -- \\

(T5 Small) &
{tiny toes and button nose, a bundle of joy soon to expose!}
& 1.00 & 1.00 & 1.00 & 1.00 \\

(Autoencoder) &
{tiny toes and button nose, a bundle of joy soon to \colorbox{red!30}{to be seen}.}
& 0.74 & 0.80 & 0.84 & 0.86 \\

(LLaMAFormer) &
{tiny toes and button nose\colorbox{red!30}{:} a bundle of joy soon to expose}
& 0.72 & 0.89 & 0.86 & 0.93 \\

({\methodname}) &
{tiny toes and button nose\colorbox{red!30}{:} a bundle of joy soon to expose!}
& 0.80 & 0.93 & 0.92 & 0.97 \\

\hline

(Input) &
{in some cases the number is 120,000,130,000.}
& -- & -- & -- & -- \\

(T5 Small) &
{in some cases the number is 120,000,130,000.}
& 1.00 & 1.00 & 1.00 & 1.00 \\

(AutoEncoder) &
{in some cases the number is 120,000,130,000.}
& 1.00 & 1.00 & 1.00 & 1.00 \\

(LLaMAFormer) &
{in some cases the number is 120,000,130,000.}
& 1.00 & 1.00 & 1.00 & 1.00 \\

({\methodname}) &
{in some cases the number is 120,000,130,000.}
& 1.00 & 1.00 & 1.00 & 1.00 \\

\hline
\end{tabular}
\end{adjustbox}
\label{tab:qualitative_res_2}
\end{table*}

\subsection{Comparison with State-of-the-Art Methods}
In this section, we compare the performance of our {\methodname} with state-of-the-art models that showcase advanced techniques in English autoencoding and language modeling. NUGGET \cite{qin2023nugget} employs a BART encoder incorporating with a feed-forward network to distill logits and extract “nuggets”—concise text segments that minimize memory usage while keeping a high compression ratio. In contrast, ICAE \cite{ge2023context}, developed by Microsoft, builds on the Llama-2-13b architecture by integrating teacher forcing and an additional 70 million parameters, generating memory slots that improve both compression efficiency and autoencoding performance. Meanwhile, Google’s T5 offers a range of scalable variants, with T5-Small (approximately 70 million parameters) delivering remarkable results despite its reduced complexity.

We scrutinize the performance of {\methodname}, which incorporates a meticulous noise process across four corpora. In autoencoding tasks, judiciously selected {\vectorname} vectors for text generation reveal that {\methodname} overtakes several transformer-based architectures in memory compression and conservation. Conventional transformers, which are especially reliant on self-attention, are hindered in their ability to narrow the bottleneck without incurring significant information loss. To substantiate this claim, we modified a Vaswani-style transformer by reducing its feed-forward layer fourfold and incorporating additional residual connections; however, this variant suffered from extreme overfitting and unendurable information loss. In contrast, while NUGGET effectively extracts succinct segments from the encoder output, it does so at almost double the parameter cost of our model. Moreover, our observations further reveal that not all encoder-produced vectors uniformly capture the complete sentence context. Additionally, the implementation of noise injection as detailed in \citep{freitag2018unsupervised} did not yield significant improvements in our experimental settings. To evaluate our data perturbation strategy and analyze {\methodname}'s performance without a denoising objective, we tested the model on the WMT19 dataset. {\methodname} achieved a BLEU score of 74.98, a BERT-Score of 96.40, and a perplexity (PPL) of 12.54. These results show a significant decline in performance compared to training with a denoising objective, indicating that {\methodname}, LLaMAFormer, and the AutoEncoder baseline benefit greatly from denoising-based learning. To evaluate the practical usability of {\methodname} and its robustness to test-time noise, we evaluated the model on the PwC and WMT19 datasets under noisy inference conditions at 50\% and 100\% Kizuki vector retention (Table \ref{tab:noisy_test}). Taking the WMT19 dataset at 100\% retention as an example, {\methodname} achieved a BLEU score of 92.57, a BERTScore of 98.58, and a perplexity (PPL) of 5.16. Introducing noise at inference time results in a marginal degradation in exact-match metrics, such as a 1.56 point drop in BLEU and a 0.31 increase in PPL relative to the clean baseline. However, deep semantic similarity remains exceptionally stable, with BERTScore degrading by less than 0.3 points across all settings. This empirically demonstrates that because {\methodname} is trained with a denoising objective, it successfully learns to look past surface-level token corruption to preserve and decompress the core semantic meaning of the text. These findings confirm the model's high resilience and practical utility in real-world scenarios where input text may be flawed. In contrast, as shown in Table \ref{tab:quantitative_res_1}, {\methodname} achieves a compression ratio of 3.86$\times$ on the PwC corpus while maintaining competitive BLEU scores. With only 86 million parameters, it outperforms T5-Small by providing a 153$\times$ reduction in parameter count compared to ICAE and a 1.88$\times$ reduction compared to NUGGET. Additionally, it invariably achieves BERT scores above 99.4 and ROUGE scores exceeding 98, despite encountering a modest BLEU drop of 1.67 to 2.47 points when compared to state-of-the-art methods. Further validation on WMT14 and BookCorpus depicted in Table \ref{tab:quantitative_res_2} demonstrates that {\methodname} excels in all evaluation metrics except for compression ratio, where our autoencoder significantly outperforms all other methods. Although the autoencoder compresses the latent space more aggressively, its perplexity is markedly higher than that of our {\methodname}, which indicates lesser confidence in the token prediction. Remarkably, by passing 20\% of the {\vectorname} vectors to the decoder’s cross-attention layer and applying LZMA entropy coding, the compression ratio on PwC improved by 1.39 points over ICAE, with similar excellent results on WMT14 (5.07), WMT19 (5.08), and BookCorpus (5.10), all with insignificant quality loss. Furthermore, our {\methodname} produces superior bits-per-character (bpc) scores—WMT14: $1.5106\times10^{-1}$, WMT19: $4.5817\times10^{-1}$, PwC: $3.5102\times10^{-1}$, and BookCorpus: $2.9544\times10^{-1}$—which are significantly lower than those of the autoencoder, thereby demonstrating more efficacious data representation and more precise token prediction. In sum, {\methodname} offers a perspicacious harmony between rigorous memory compression and high-fidelity text generation, rendering it an invaluable asset for various downstream tasks.

In parallel, we examined our Autoencoder variant that reduces the initial dimensionality into a fixed-size latent space, which is then further compressed using general-purpose lossless compressors such as LZMA \citep{ziv1978compression}, GZIP \citep{deutsch1996gzip}, ZLIB \citep{deutsch1996zlib}, and ZSTD \citep{zstdcite}. Our rigorous tests revealed that LZMA consistently achieved the highest compression ratios—67$\times$ on PwC, 33$\times$ on both WMT19 and WMT14, and 25$\times$ on BookCorpus. Moreover, it surpasses transformer-based lossless compressors such as GPT-AC \citep{huang2023approximating} and TRACE \citep{mao2022trace} in compression efficiency on the BookCorpus dataset, while exhibiting only minor degradation in text quality for sequences where $\ell<$ 1000. This approach significantly optimizes memory usage, reducing storage requirements by 32GB to 44GB per epoch across datasets. Notably, the Autoencoder is 196$\times$ smaller than the best-performing transformer-based alternative while incurring only a 4\% disparity in quality performance. The empirical findings demonstrate that the Autoencoder surpasses NN-based lossless compressors in terms of memory efficiency. Building on these insights, our lightweight neural network variant, LLaMAFormer, achieves prominent performance over the Autoencoder across all datasets and surpasses {\methodname} on a couple of metrics in the WMT19 dataset while operating with a minimal 50M parameters. This demonstrates that a streamlined transformer-based architecture can effectively process injected noise to denoise text and generate high-quality outputs.

Our extensive qualitative experiments present three complementary approaches to efficient text processing. {\methodname} achieves a 5.39$\times$ compression ratio while maintaining competitive BLEU scores through {\vectorname} vectors, requiring only 86M parameters. The LLaMAFormer variant, incorporating LLaMA architectural components, further reduces the parameter count to 50M while maintaining strong performance, particularly excelling on the WMT19 dataset. Finally, our Autoencoder approach demonstrates exceptional compression capabilities, achieving up to 67$\times$ compression ratios on the PWC dataset and significant memory savings across all tested corpora. Together, these approaches showcase different strategies for balancing model efficiency with generation quality, offering practical solutions for resource-constrained NLP applications.

\subsection{Qualitative Results}
The qualitative performance of ICAE, T5 Small, Autoencoder, LLaMAFormer, and TextEconomizer has been depicted in Table \ref{tab:qualitative_res}, effectively showcasing the superiority of our Transformer-based TextEconomizer for the lossy autoencoding task. The examples provided in the table encompass complex sentences, numerical data and special characters, interrogatives, negations, proper nouns, dates, technical terms, idiomatic expressions, wordplay, and rhyme. The narrative of sentences progresses seamlessly from anticipation to connection, then transition, and finally pinnacles in an unexpected twist, thereby highlighting the robustness of our qualitative analysis.
In particular, the results reveal that T5 Small struggles with punctuation and long sentences, whereas ICAE often substitutes words with synonyms and occasionally generates redundant sequences that, while semantically uniform, can be unnecessarily lengthy. In contrast, the Autoencoder exhibits punctuation challenges, particularly at sentence boundaries, yet unfailingly maintains the intended meaning and avoids incoherent expressions. For instance,  as seen in the fourth example, it tends to replace double quotes with single quotes and sometimes redundantly repeats words (e.g., ``investment investment”). LLaMAFormer, on the other hand, occasionally introduces spaces within email addresses and faces difficulties when reconstructing lengthy names of people. The 50\% vector reduction variant tends to insert extraneous punctuation after years (e.g., converting ``2017" to ``2017:") and sometimes expands contractions (e.g., ``they've" to ``they have"), while the 20\% vector reduction variant may replace words but not inappropriate and omit single quotes surrounded within double quotes.

Our proposed model, TextEconomizer, occasionally omits punctuation in quotes or within sentences containing dense punctuation, yet it consistently maintains the core semantic content—a critical factor in lossy text compression. Additionally, TextEconomizer sometimes skips apostrophes in contractions (e.g., rendering ``she would" as ``she’d" without the apostrophe), yet the 50\% {\vectorname} vector variant exhibits improved accuracy in tackling punctuation within quotations, albeit with minor punctuation substitutions. With 20\% {\vectorname} vector variant, nevertheless, misses single-digit numbers in lengthy sentences and, in cases with consecutive commas, alters base-form verbs to past participles rarely. 

Overall, these qualitative remarks suggest that our framework is well-suited for any optimized encoder-decoder neural network, particularly transformer-based models. Specifically, our proposed TextEconomizer not only produces fine-grained outputs with exceptional memory efficiency compared to baseline models but also effectively mitigates diverse textual errors through a meticulous noise injection process. This design choice clarifies occasional punctuation mistakes, which are overshadowed by the model’s ability to maintain the sentence’s core meaning; signifying its achievement in lossy text compression. Moreover, all methods, including T5 Small, reveal proportional efficacy when processing shorter sentences, as evidenced by the final example.

\section{Conclusion and Future Work}
\label{conclusion_futureWork}
In this study, we present an encoder-decoder framework and a memory-efficient baseline for the task at hand by proposing TextEconomizer; a monolingual transformer that leverages Kizuki vectors and a novel text noising strategy. TextEconomizer refines the latent representation by strategically filtering its most informative vectors and by using entropy coding algorithms to condense the latent space more efficiently, effectively addressing the complex linguistic intricacies inherent in the task. TextEconomizer outperforms transformer-based baseline methods in terms of parameter and memory efficiency across various corpora, with only a negligible $\approx$2\% reduction in N-gram precision compared to the best-performing models. Moreover, the adaptability of our framework enables seamless integration with any encoder-decoder network. We took advantage of this flexibility by incorporating LLaMAFormer, a remarkably memory-efficient transformer variant that demonstrates superior text generation performance compared to our autoencoder. In addition, we integrated a state-of-the-art memory-efficient autoencoder with our sophisticated noisy text processing approach, thereby challenging the notion that autoencoder-based methods are solely adequate for image compression and extending their relevance to text-based tasks. Our work opens new avenues for efficient natural language processing in resource-constrained settings. Future research directions include knowledge distillation from multilingual models to enhance our monolingual model, quantization for improved compression outcomes, text compression in low-resource languages such as Bangla, and large-scale experiments integrating contrastive learning techniques.

\section{Acknowledgments}
This research was partly funded by the ICT division of the Government of the People’s Republic of Bangladesh for the 2024-25 financial year (tracking no: 26FS110481).

\bibliographystyle{elsarticle-harv} 
\setcitestyle{authoryear,open={(},close={)}}
\bibliography{reference}






\end{document}